\documentclass[10pt,twocolumn,letterpaper]{article}

\usepackage{cvpr}              

\usepackage{graphicx}
\usepackage{amsmath}
\usepackage{amssymb}
\usepackage{booktabs}
\usepackage{tablefootnote} 
\usepackage{adjustbox}
\usepackage{multicol}
\usepackage{multirow}
\usepackage{color, colortbl}

\usepackage{cuted}
\usepackage{afterpage}

\newcommand{\etcno}{\textit{etc}}

\newcommand{\ieno}{\textit{i}.\textit{e}.}
\newcommand{\egno}{\textit{e}.\textit{g}.}
\newcommand{\ours}{TaskRes}

\definecolor{Gray}{gray}{0.9}

\newcommand{\tcb}{}
\newcommand{\tcr}{}

\usepackage[pagebackref,breaklinks,colorlinks]{hyperref}

\usepackage[capitalize]{cleveref}
\crefname{section}{Sec.}{Secs.}
\Crefname{section}{Section}{Sections}
\Crefname{table}{Table}{Tables}
\crefname{table}{Tab.}{Tabs.}

\def\cvprPaperID{2643} 
\def\confName{CVPR}
\def\confYear{2023}

\begin{document}

\title{Task Residual for Tuning Vision-Language Models}


\author{Tao Yu$^{1,2}$\thanks{Equal contribution.}~~~~Zhihe Lu$^{1*}$~~~Xin Jin$^{1,2}$~~~Zhibo Chen$^2$~~~Xinchao Wang$^{1}$\thanks{Corresponding author.}\\
$^1$National University of Singapore ~$^2$University of Science and Technology of China \\
{\tt\small yutao666@mail.ustc.edu.cn, zhihelu@nus.edu.sg, jinxustc@mail.ustc.edu.cn,} \\ {\tt\small chenzhibo@ustc.edu.cn, xinchao@nus.edu.sg}}

\maketitle

\begin{abstract}
Large-scale vision-language models (VLMs) pre-trained on billion-level data have learned general visual representations and broad visual concepts. In principle, the well-learned knowledge structure of the VLMs should be inherited appropriately when being transferred to downstream tasks with limited data.
However, most existing efficient transfer learning (ETL) approaches for VLMs either damage or are excessively biased towards the prior knowledge, \egno, prompt tuning (PT) discards the pre-trained text-based classifier and builds a new one while adapter-style tuning (AT) fully relies on the pre-trained features.
To address this, we propose a new efficient tuning approach for VLMs named Task Residual Tuning  (\ours), which performs directly on the text-based classifier and explicitly decouples the prior knowledge of the pre-trained models and new knowledge regarding a target task. Specifically, \ours~keeps the original classifier weights from the VLMs frozen and obtains a new classifier for the target task by tuning a set of prior-independent parameters as a residual to the original one, which enables reliable prior knowledge preservation and flexible task-specific knowledge exploration.
The proposed \ours~is simple yet effective, which significantly outperforms previous ETL methods (\egno, PT and AT) on 11 benchmark datasets while requiring minimal effort for the implementation. Our code is available at \href{https://github.com/geekyutao/TaskRes}{https://github.com/geekyutao/TaskRes}.
\end{abstract}

\vspace{-8pt}
\section{Introduction}
\label{sec:intro}

\begin{figure}[t]
	\begin{center}
		\includegraphics[scale=0.50]{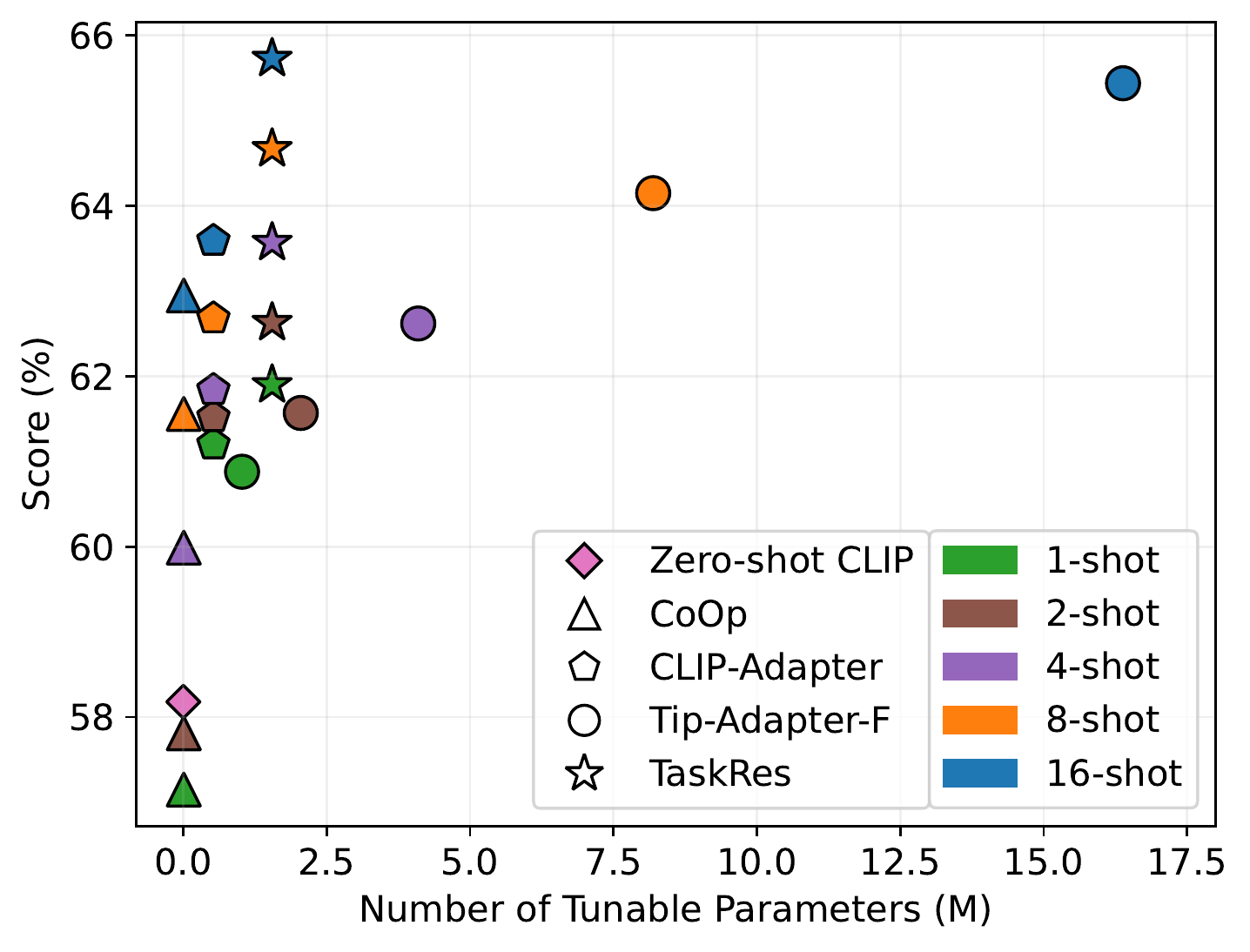} 
	\end{center}
        \vspace{-12pt}
	\caption{Performance comparison between Zero-shot CLIP \cite{radford2021learning}, CoOp \cite{zhou2022learning}, CLIP-Adapter \cite{gao2021clip}, Tip-Adapter-F \cite{zhang2022tip} and our \ours~on ImageNet with few shot settings.}
	\label{fig:params_acc}
	\vspace{-12pt}
\end{figure}

Over the past decade, deep learning-based visual recognition models \cite{krizhevsky2017imagenet,simonyan2014very,szegedy2015going,he2016deep,dosovitskiy2020image} have achieved great success. These state-of-the-art models are often trained on a large amount of image and discrete label pairs. The discrete label is generated by converting a detailed textual description, \egno, ``American curl cat'', into a simple scalar, which strikingly eases the computation of loss function. However, this also results in two evident limitations: (i) the rich semantics in the textual description are underused, and (ii) the trained models are limited to recognizing the close-set classes only.

\begin{figure*}[t]
	\begin{center}
		\includegraphics[scale=0.42]{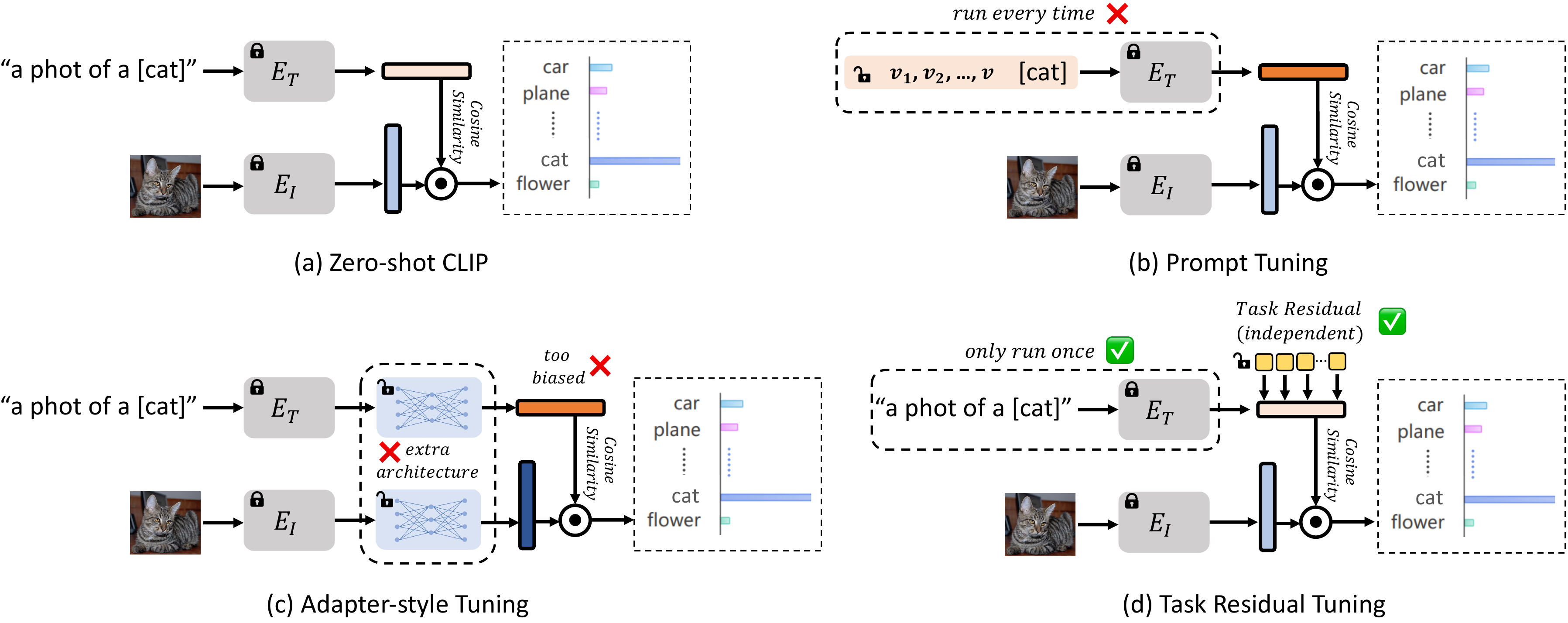} 
	\end{center}
        \vspace{-10pt}        
	\caption{Illustration of (a) Zero-shot CLIP, (b) prompt tuning, (c) adapter-style tuning and (d) our proposed Task Residual Tuning (\ours). Our method introduces a \textit{prior-independent} task residual to the fixed pre-trained classifier (\ieno, text embeddings of CLIP), being free of running the text encoder every time or extra architecture design.}
	\label{fig:motivation}
	\vspace{-11pt} 
\end{figure*}

Recent large-scale vision-language model (VLM) pre-training \tcb{\cite{radford2021learning,jia2021scaling,yuan2021florence,li2021align,alayrac2022flamingo}} eliminates those limitations by learning visual representations via textual supervision.
For instance, texts and images are encoded and mapped into a unified space via a contrastive loss during pre-training \cite{radford2021learning}. The pre-trained text encoder can then be used to synthesize a text-based classifier for image recognition given the corresponding natural language descriptions as shown in Figure \ref{fig:motivation} (a). Those pre-trained VLMs have demonstrated a powerful transferability on a variety of downstream tasks in a zero-shot manner.
However, the effectiveness of the aforementioned models heavily relies on their large-scale architectures and training datasets. For instance, CLIP \cite{radford2021learning} has up to 428 million parameters and is trained on 0.4 billion text-image pairs, while Flamingo \cite{alayrac2022flamingo} boasts up to 80 billion parameters and is trained on a staggering 2.1 billion pairs.
This makes it impractical to fully fine-tune the model on downstream tasks in a low-data regime.

For that reason, efficient transfer learning (ETL) \cite{zhou2022learning,zhou2022conditional,gao2021clip,zhang2022tip} on pre-trained VLMs has gained popularity. ETL represents transfer learning to downstream tasks in both parameter- and data-efficient manner. The core of ETL is twofold: (i) properly inheriting the well-learned knowledge structure of VLMs, which is already transferable; (ii) effectively exploring the task-specific knowledge given limited data. However, most existing ETL approaches, \ieno, prompt tuning (PT) \cite{zhou2022learning,zhou2022conditional} and adapter-style tuning (AT) \cite{gao2021clip,zhang2022tip}, either damage the prior knowledge of VLMs or learn the new knowledge of a task in an inappropriate/insufficient way. For example, instead of using the pre-trained text-based classifier, CoOp \cite{zhou2022learning} (in Figure \ref{fig:motivation} (b)) is proposed to learn a continuous prompt for synthesizing a completely new one, which inevitably causes the loss of previous knowledge. Consequently, CoOp underperforms Zero-shot CLIP by 1.03\%/0.37\% in 1-/2-shot learning on ImageNet (see Figure \ref{fig:params_acc}). In contrast, CLIP-Adapter \cite{gao2021clip} preserves the pre-trained classifier, but is excessively biased towards the prior knowledge when learning a new task, \ieno, it transforms the pre-trained classifier weights to be task-specific as illustrated in Figure \ref{fig:motivation} (c). This results in an inferior new knowledge exploration, thereby a lower accuracy as shown in Figure \ref{fig:params_acc}.

For better ETL on pre-trained VLMs, we propose a new efficient tuning approach named Task Residual Tuning (\ours), which performs directly on the text-based classifier and explicitly decouples the old knowledge of the pre-trained models and the new knowledge for a target task.
The rationale is that the decoupling enables a \textit{better old knowledge inheritance from VLMs} and a \textit{more flexible task-specific knowledge exploration}, \ieno, the learned knowledge \textit{w.r.t.} the task is \textit{independent} on the old knowledge.
Specifically, \ours~keeps the original classifier weights frozen and introduces a set of \textit{prior-independent} parameters that are added to the weights. These additive parameters, tuned for adaptation to the target task, are thus named ``task residual''.

To gain insight into how \ours~works, we perform extensive experiments across 11 benchmark datasets \cite{zhou2022learning} and conduct a systematic investigation of learned task residuals. The experimental results demonstrate that
\tcb{introducing task residual can significantly enhance}
the transfer performance. We visualize the correlation between the magnitude of learned task residual and the difficulty of transferring a pre-trained model to a downstream task, and observe that the magnitude increases with the transfer difficulty. This suggests the residual is automatically adapted to the task to fully explore the new knowledge, thereby achieving a new state-of-the-art performance on 11 diverse datasets. Furthermore, it is worth noting that our method requires minimal effort for the implementation, \ieno, technically adding one line of code only.
Our contributions are summarized below:
\begin{itemize}
    \item We for the first time emphasize the necessity of a proper knowledge inheritance from pre-trained VLMs to downstream tasks via ETL, reveal the pitfalls of existing tuning paradigms, and conduct an in-depth analysis to manifest that decoupling the old pre-trained knowledge and the new task-specific knowledge is the key.
    \item We propose a new efficient tuning approach named Task Residual Tuning (\ours), which achieves a better old knowledge inheritance from VLMs and a more flexible task-specific knowledge exploration.
    \item \ours~is convenient for use, which needs a few tuning parameters and effortless implementation.
\end{itemize}





\section{Related Work}
\subsection{Vision-Language Models}
We mainly review the literature of vision-language models (VLMs) on language-driven visual representation learning (LDVRL) \cite{socher2013zero,frome2013devise,elhoseiny2013write,lei2015predicting,joulin2016learning,gomez2017self,li2017learning,desai2021virtex,sariyildiz2020learning,radford2021learning,anderson2018bottom}. The core of LDVRL is to map texts (language) and images into a common space so that the text representations can be used for visual classification. To achieve this, two encoders for texts and images respectively, and specific loss functions as regularization are needed. 

Early works explore unsupervised pre-trained models \cite{socher2013zero} or the skip-gram text modeling \cite{frome2013devise,mikolov2013efficient,mikolov2013distributed} for text embedding while sparse coding and vector quantization \cite{socher2013zero,coates2011importance} or Classeme features \cite{elhoseiny2013write,torresani2010efficient} for visual encoding. For training objectives, MSE \cite{socher2013zero}, self-supervised topic probabilities matching \cite{gomez2017self} or multi-class logistic loss are adopted. However, those methods are limited to small datasets \cite{joulin2016learning,li2017learning,desai2021virtex} and weakly representative encoding backbones \cite{socher2013zero,coates2011importance}, which tremendously hinders their transferability. 

In contrast, recent works \cite{radford2021learning,jia2021scaling,li2021supervision} have advanced previous ones by leveraging billion-level data of image-text pairs from internet and super powerful neural networks, \egno, Transformers \cite{vaswani2017attention,dosovitskiy2020image,touvron2021training,yuan2021tokens,yu2022metaformer}, for representation learning. By end-to-end pre-training with a contrastive loss, large-scale VLMs demonstrate remarkable transferability on various downstream tasks in a zero-shot evaluation manner. Standing on the strong transfer capability of large-scale vision-language pre-training, we further explore its potential in efficient transfer learning.

\subsection{Efficient Transfer Learning}
Pre-training a neural network on large-scale datasets, \egno, ImageNet \cite{deng2009imagenet} and WebImageText \cite{radford2021learning}, then fine-tuning it on downstream tasks has been a common step of transfer learning \cite{yang2022deep,yang2022KF,Ye2023CVPR,lian2022scaling,LiuSonghua2023CVPR,liu2022dataset}.
Here, we mainly focus on efficient transfer learning (ETL) on pre-trained VLMs. ETL represents both parameter- and data-efficient transfer learning, in which a small number of parameters are tuned and a small set of data is utilized. Existing works on ETL can be grouped into two categories: prompt tuning \cite{zhou2022learning,zhou2022conditional,lu2022prompt} and adapter-style tuning (AT) \cite{gao2021clip,zhang2022tip}. Specifically, prompt engineering \cite{jiang2020can,shin2020autoprompt} is first explored to generate appropriate discrete prompts for downstream tasks. Later, learning continuous prompts \cite{zhou2022learning,lu2017knowing} that can better adapt to a new task shows a more advanced performance. However, these methods face two issues: (i) they need the pre-trained text encoder to engage in the whole training, limiting its scalability and increasing the computational overhead; (ii) they abandon the well pre-trained text-based classifier and generate a new one, which causes the loss of prior knowledge of VLMs.

AT-based methods \cite{gao2021clip,zhang2022tip} solve the above issues by using the text encoder once for text-based classifier generation and then focusing on adapting text/image features only. This straightforward design can achieve even better performance, but it heavily relies on the prior knowledge/pre-trained features, which causes an inferior new knowledge exploration. To address this, we propose a new efficient tuning approach, Task Residual Tuning (TaskRes), for a better and more flexible task-specific knowledge learning by a \textit{prior-independent} ``task residual''.

\subsection{Few-Shot Learning}
Few-shot learning (FSL) aims to adapt a model to novel tasks/classes with a handful of labeled examples. Conventional FSL methods \cite{schmidhuber1987evolutionary,lu2021simpler,cai2018memory,finn2017model,lu2022prediction,goldblum2020unraveling} often meta-learn on abundant data from base classes for adaptive capability. However, the need of learning on base datasets restricts their scalability. Recent VLM pre-training works \cite{radford2021learning,jia2021scaling} offer an effective alternative, which does not require a base dataset. They show that the pre-trained models can already achieve astonishing performance on many downstream tasks in a zero-shot manner. ETL \cite{zhou2022learning,gao2021clip} can further boost the performance. In this work, we propose a novel ETL method for adapting VLMs to downstream tasks and evaluate its effectiveness on few-shot tasks.


\section{Preliminaries}
We briefly introduce the adopted VLM, \ieno, contrastive language-image pre-training (CLIP) \cite{radford2021learning}, and recap two mainstream approaches for ETL on VLMs, \ieno, prompt tuning and adapter-style tuning.

\subsection{Contrastive Language-Image Pre-training}
The CLIP model \cite{radford2021learning} is designed to obtain visual representations through natural language supervision.
It is trained on 0.4 billion image-text pairs, where image features from an image encoder and text features from a text encoder are aligned within a unified embedding space using a contrastive learning loss, allowing CLIP to effectively capture broad visual concepts and learn general visual representations. 
When testing, CLIP can classify a query image into $K$ possible categories.
This is achieved by calculating the cosine similarity between a query image embedding $\mathbf{z}$, obtained from the image encoder with projection, and the text embeddings $\left \{ \mathbf{t}_i  \right \} _{i=1}^{K}$, which are derived by inputting texts (\egno, ``a photo of a \{class\}") into the text branch.
The predicted probability for class $i$ is formulated as
\begin{equation}
    p(y=i|\mathbf{z})=\frac{\mathrm{exp}(\mathrm{sim}(\mathbf{z},\mathbf{t}_i)/\tau)}{\sum_{j=1}^{K}\mathrm{exp}(\mathrm{sim}(\mathbf{z},\mathbf{t}_j)/\tau)},
\end{equation}
where $\mathrm{sim}(\cdot,\cdot)$ indicates cosine similarity and $\tau$ is the learned temperature of CLIP.

\subsection{Revisiting Previous Tuning Paradigms}

Inspired by the success of ETL approaches in natural language processing, \egno, prompt tuning \cite{li-liang-2021-prefix,lester-etal-2021-power} and Adapter \cite{houlsby2019parameter}, recent advances (\egno, CoOp \cite{zhou2022learning} and CLIP-Adapter \cite{gao2021clip}) borrow their ideas to ETL on VLMs.

CoOp, for the first time, introduces prompt tuning to VLMs. Instead of using fixed text prompt contexts such as ``a photo of a", CoOp proposes to use $M$ learnable context vectors $\{\mathbf{v}_m\}_{m=1}^{M}$ as the task-specific templates.
The prompt given to the text encoder then becomes $\{\mathbf{v}_1, \mathbf{v}_2, \cdots, \mathbf{v}_M, \mathbf{c}_i\}$ \tcb{where $\mathbf{c}_i$ is the embedding of class $i$}.
During the whole training process, CoOp keeps the parameters of pre-trained VLMs frozen and only tunes the learnable vectors $\{\mathbf{v}_m\}_{m=1}^{M}$.

Adapter-style tuning introduces additional modules $\phi_{\omega}(\cdot)$ with tunable parameters $\omega$ to the pre-trained models to transform the pre-trained features $\mathbf{f}$ to new ones $\mathbf{f}'$.
In general, adapter-style tuning can be formulated as
\begin{equation}
    \mathbf{f}'=\mathbf{f} + \alpha \phi_{\omega}(\mathbf{f}),
\end{equation}
where $\alpha$ is a scaling factor. In CLIP-Adapter, the adapter module $\phi_{\omega}$ is composed of two linear transformation layers and a ReLU activation between them. CLIP-Adapter investigates both visual and text adapters, \ieno, applying the adapter module to the image and text branches of CLIP, respectively, and shows that they have comparable performance.
During training on downstream tasks, adapter-style methods only tune their adapter modules.

\section{Approach}
In this section, we start by identifying the pitfalls of the existing ETL paradigms on VLMs. Those pitfalls motivate us to propose a new ETL approach for VLMs, named \textit{Task Residual Tuning} (\ours). The proposed \ours~is simple yet can effectively avoid the pitfalls.
\subsection{Pitfalls of Existing ETL Paradigms on VLMs}
We rethink the usage of the prior knowledge from pre-trained VLMs and the acquisition of the new knowledge regarding downstream tasks. On the one hand, large-scale VLMs trained with a vast amount of data have learned broad visual concepts that are general for a wide range of downstream vision tasks, enabling homogenization \cite{bommasani2021opportunities}. The prior knowledge should be well preserved when performing a transfer. On the other hand, despite the huge data used in the pre-training, inevitably, there are domain shifts or uncertain concepts in downstream tasks. New knowledge specific to the downstream tasks should be appropriately supplemented to the prior knowledge. However, existing ETL paradigms do not well consider the above principles and have the following two issues.

\vspace{-10pt}
\paragraph{Pitfall 1: Lack of guarantees of prior knowledge preservation in prompt tuning.} While the weights of pre-trained text branch modules (\egno, text encoder and projection) are frozen in the prompt tuning paradigm, the original well-learned classification boundary is more or less damaged. This is because the tuning of input prompts ends up with a new boundary that may forget the old knowledge without explicit regularization.
As a result, the performance of prompt tuning is limited. For instance, the performance of CoOp \cite{zhou2022learning} is not as good as Zero-shot CLIP in 1-/2-shot learning on ImageNet as shown in Figure \ref{fig:params_acc}.

\vspace{-10pt}
\paragraph{Pitfall 2: Limited flexibility of new knowledge exploration in adapter-style tuning.} The data distributions in downstream tasks often shift from the pre-training distribution, and some task-specific or fine-grained visual concepts/representations may not be well learned by pre-trained VLMs, \egno, from CLIP \cite{radford2021learning} to satellite image dataset EuroSAT \cite{helber2019eurosat}. Thus, the new knowledge regarding downstream tasks needs to be appropriately explored. We observe that adapter-style tuning may not sufficiently explore the task-specific knowledge as the input of the adapter is strictly limited to old/pre-trained features, as shown in Figure \ref{fig:motivation} (c). \textit{Regardless of whether the pre-trained features are suitable for the task, the results of the adapter depend only on them, making adapter-style tuning have limited flexibility for learning new knowledge}.

\subsection{Task Residual Tuning}
Given that the existing ETL paradigms face the aforementioned issues, we propose Task Residual Tuning (\ours) to address the issues in a simple way. 
\ours~explicitly decouples the maintenance of the old knowledge from the pre-trained VLMs and the learning of task-specific knowledge that is not overly biased on the pre-trained features. We elaborate our \ours~below.

\vspace{-10pt}
\paragraph{Fixed base classifier.} As illustrated in Figure \ref{fig:motivation} (d), our \ours~performs the tuning directly on the text-based classifier (\ieno, text embeddings)\footnote{While we can tune on image embeddings, it may encounter overfitting since the testing image embeddings are different from the training ones. Tuning on text embeddings is free of the issue. Besides, large diversity between image embeddings is not conducive to learning stable task-level information.}. \textit{Base classifier} is the text embeddings of a pre-trained vision-language model, \egno, CLIP. We denote the base classifier as $\mathbf{t} \in \boldsymbol{\mathbb{R}}^{K \times D}$ where $K$ is the number of categories and $D$ is the feature dimension. We keep the base classifier weights frozen to explicitly prevent the base classifier from being damaged.
\vspace{-10pt}
\paragraph{Prior-independent task residual.} To learn task-specific knowledge without being restricted by prior knowledge, we propose \textit{task residual}, which is a set of tunable parameters $\mathbf{x} \in \boldsymbol{\mathbb{R}}^{K \times D}$ that are not dependent on the base classifier. 
Our task residual is scaled by a factor $\alpha$ and element-wisely added to the base classifier to formulate a new classifier 
$\mathbf{t}'$ for the target task, written as
\vspace{-5pt}
\begin{equation}
    \mathbf{t}'=\mathbf{t}+\alpha\mathbf{x}.
    \label{eq:target cls}
\end{equation}
\vspace{-30pt}
\paragraph{Tuning for downstream tasks.} During tuning, we only tune the prior-independent task residual while keeping the base classifier (together with the image branch) fixed, enabling reliable old knowledge preservation and flexible new knowledge exploration. Given an image, the fixed image branch of CLIP extracts its embeddings $\mathbf{z} \in \boldsymbol{\mathbb{R}}^{D}$. The predicted probability for class $i$ is then calculated as
\begin{equation}
    p(y=i|\mathbf{z})=\frac{\mathrm{exp}(\mathrm{sim}(\mathbf{z},\mathbf{t}'_i)/\tau)}{\sum_{j=1}^{K}\mathrm{exp}(\mathrm{sim}(\mathbf{z},\mathbf{t}'_j)/\tau)}.
\end{equation}
Based on the predicted probabilities, the downstream task loss (\egno, cross-entropy loss) only updates the task residual through standard backpropagation.


\begin{figure*}[ht]
\centering
\begin{adjustbox}{minipage=\textwidth,scale=0.88}
\begin{subfigure}{0.33\textwidth}
    \includegraphics[width=\textwidth]{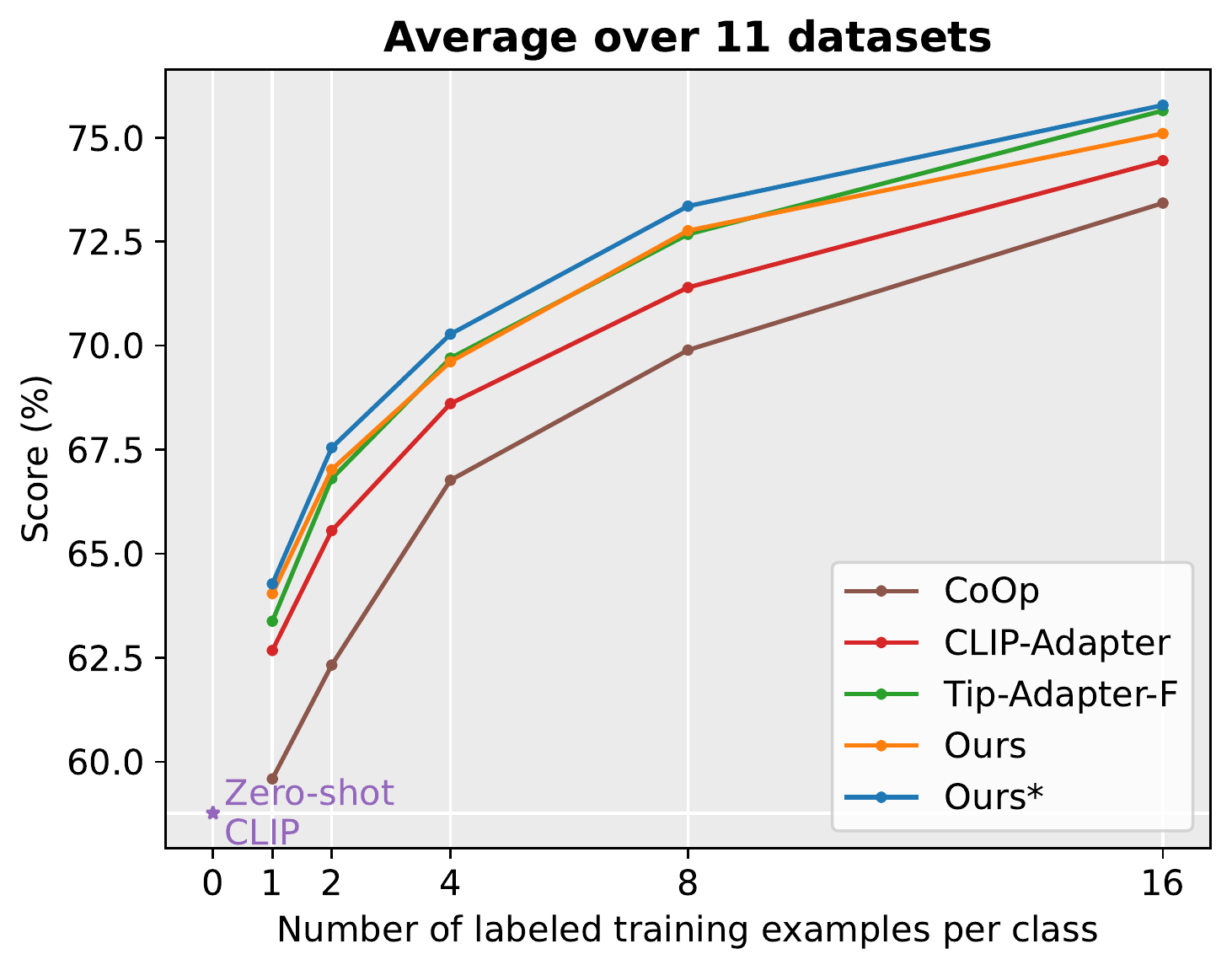}
    \vspace{-10pt}
\end{subfigure}
\hfill
\begin{subfigure}{0.33\textwidth}
    \includegraphics[width=\textwidth]{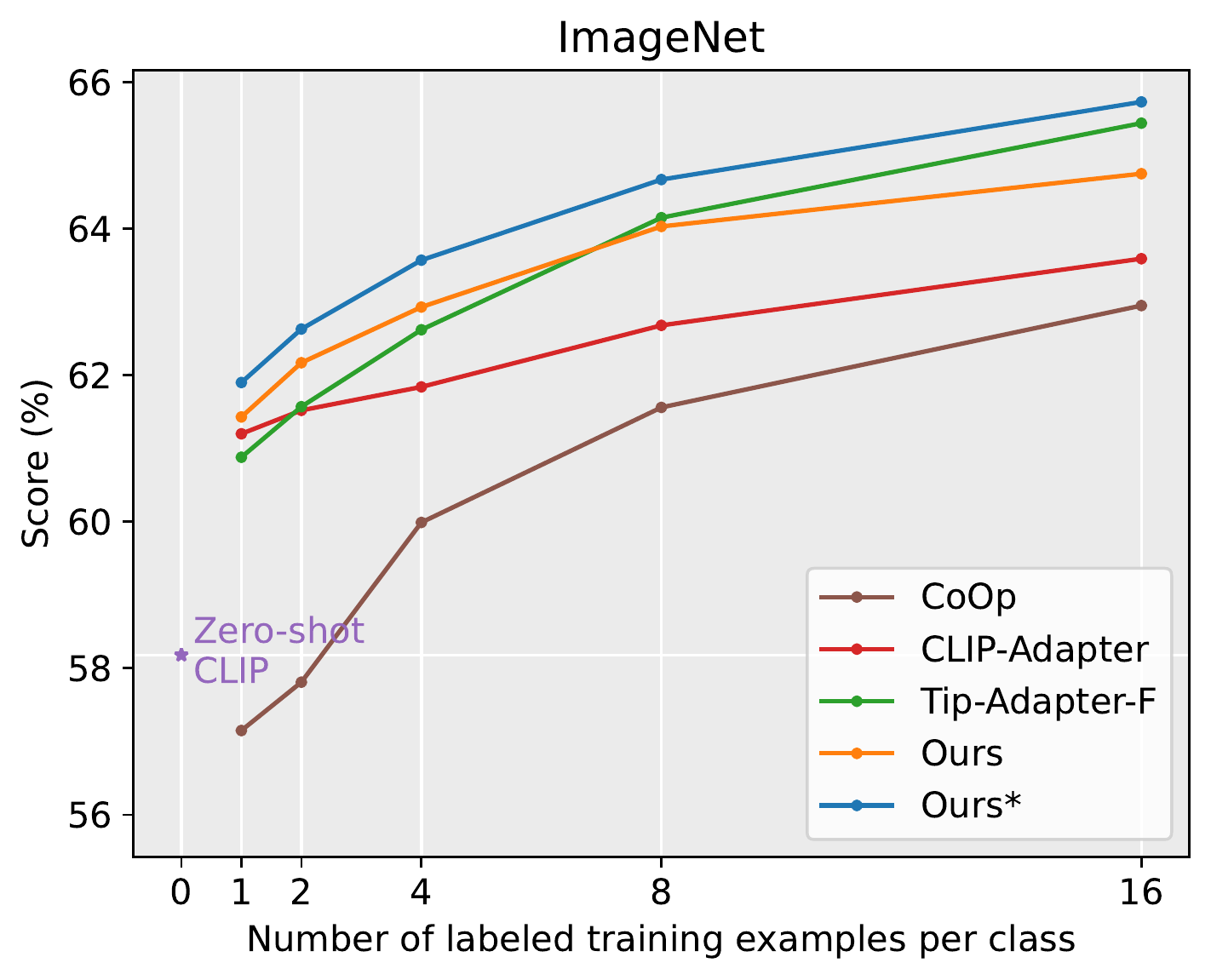}
    \vspace{-10pt}
\end{subfigure}
\hfill
\begin{subfigure}{0.33\textwidth}
    \includegraphics[width=\textwidth]{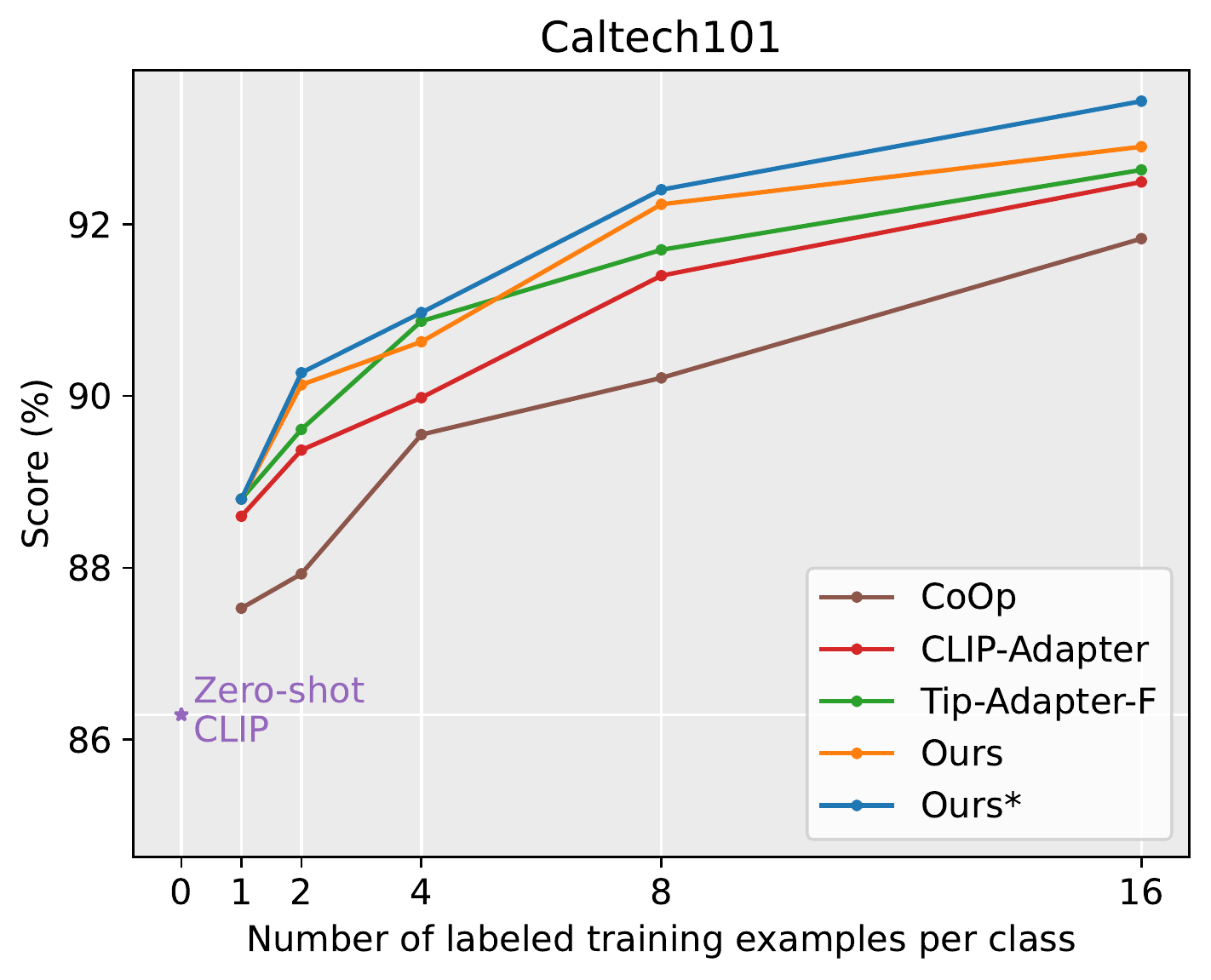}
    \vspace{-10pt}
\end{subfigure}
\begin{subfigure}{0.33\textwidth}
    \includegraphics[width=\textwidth]{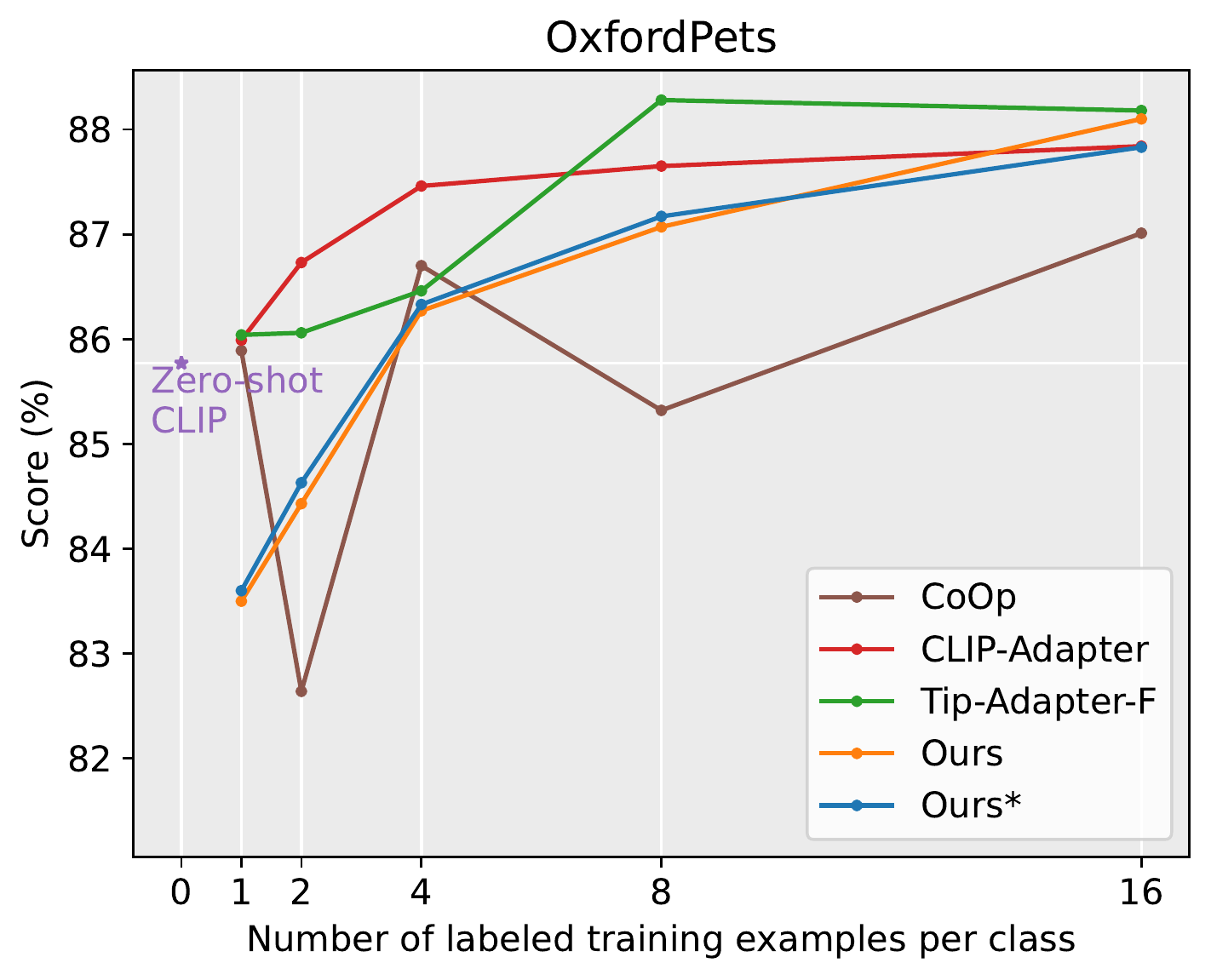}
    \vspace{-10pt}
\end{subfigure}
\hfill
\begin{subfigure}{0.33\textwidth}
    \includegraphics[width=\textwidth]{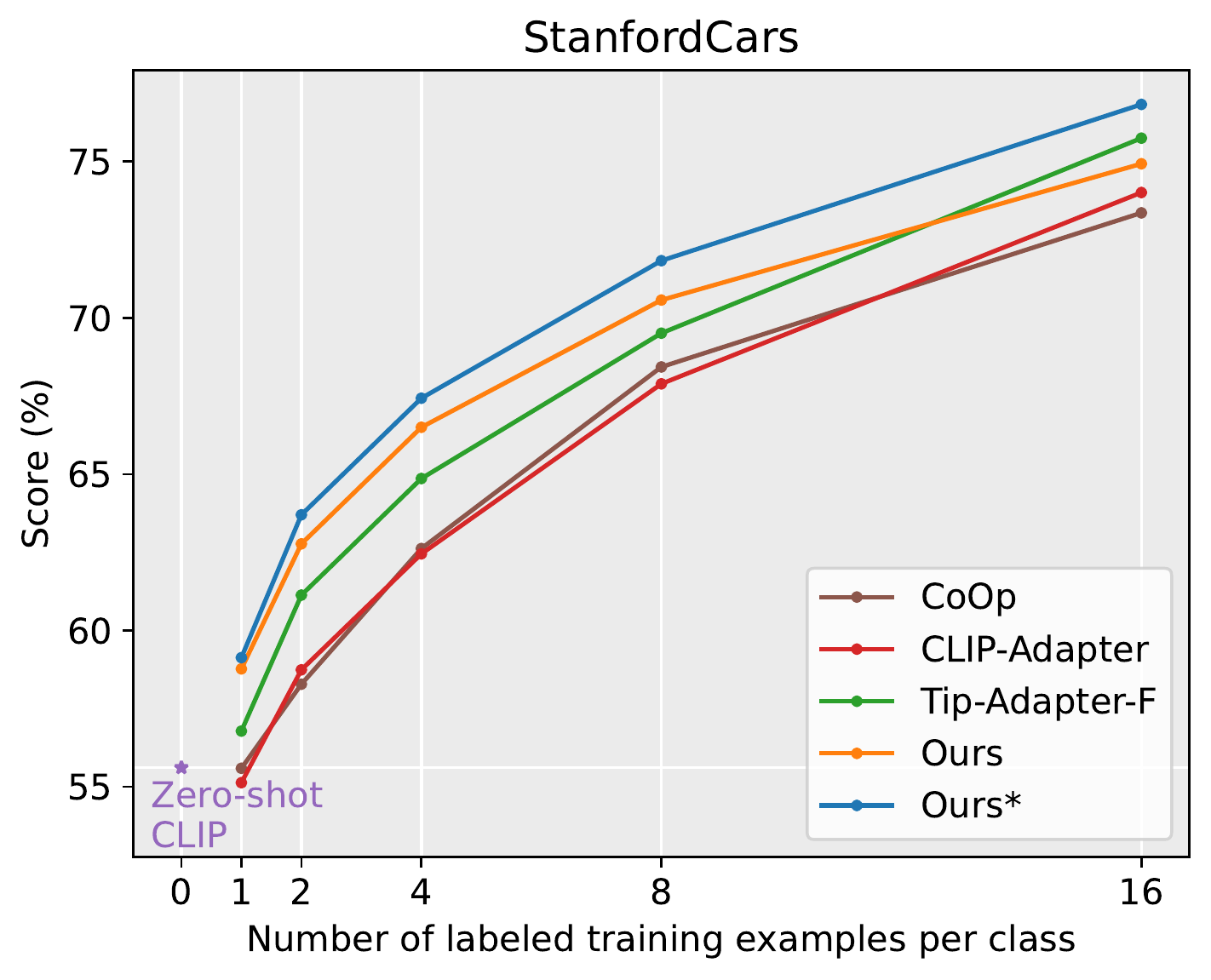}
    \vspace{-10pt}
\end{subfigure}
\hfill
\begin{subfigure}{0.33\textwidth}
    \includegraphics[width=\textwidth]{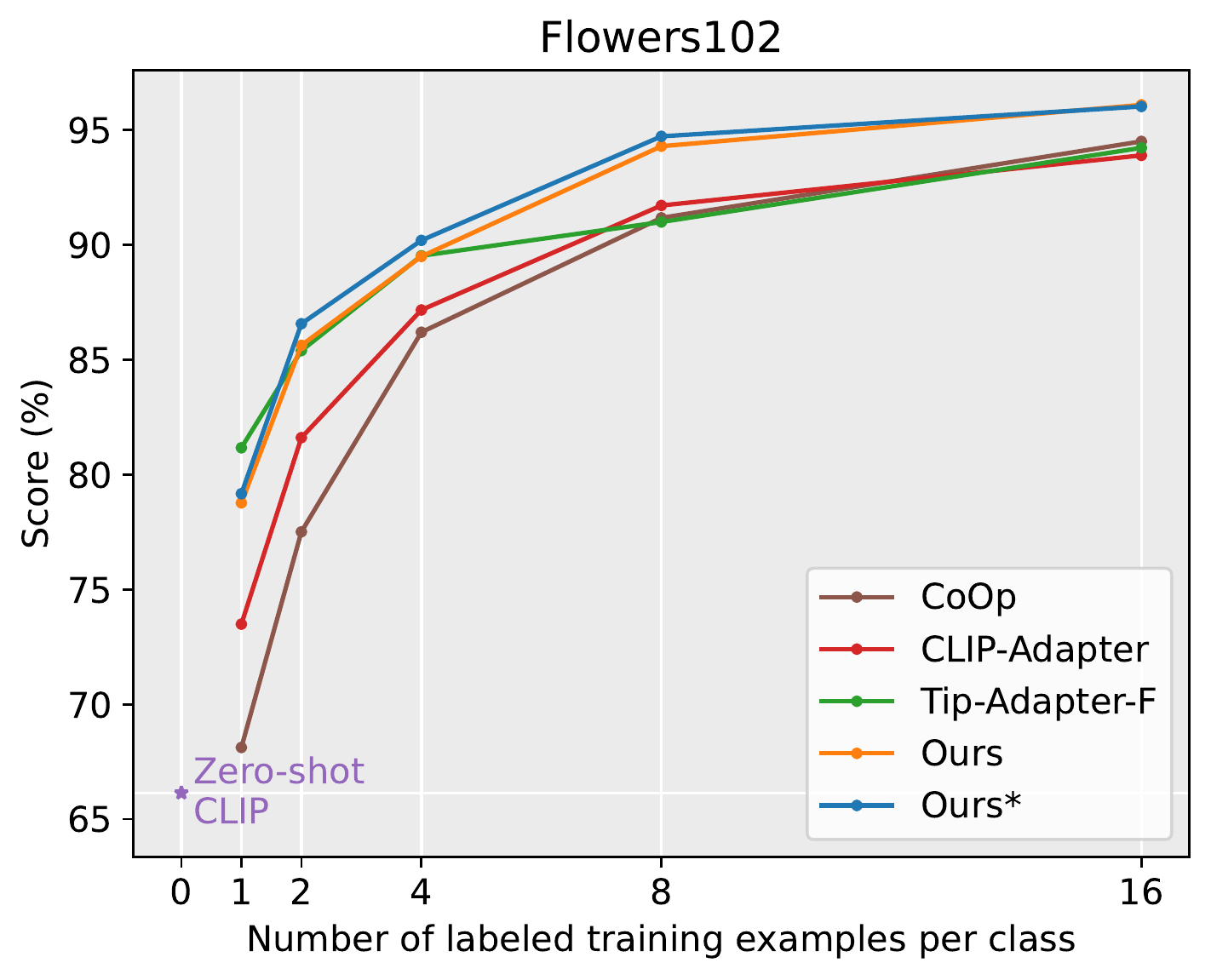}
    \vspace{-10pt}
\end{subfigure}

\begin{subfigure}{0.33\textwidth}
    \includegraphics[width=\textwidth]{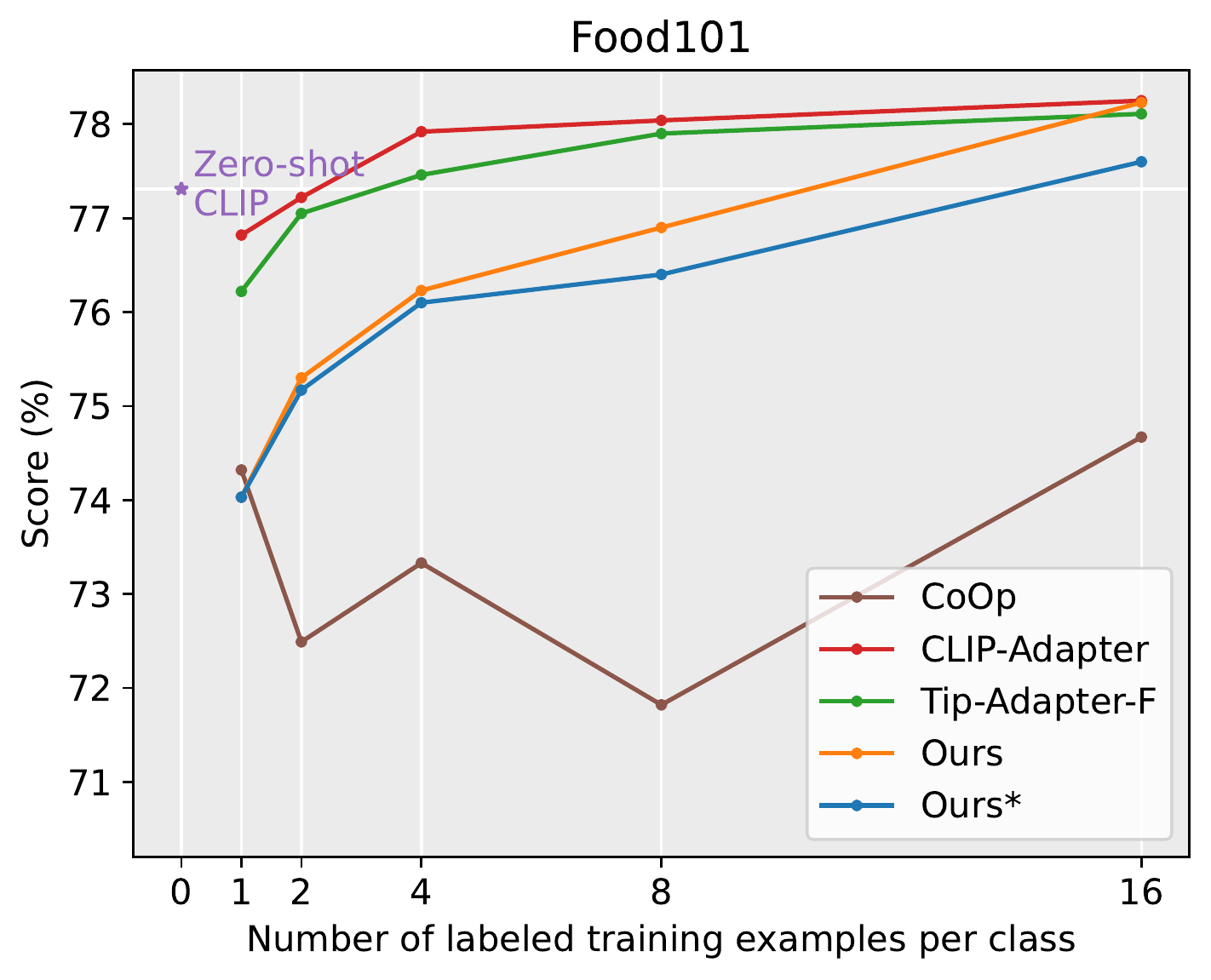}
    \vspace{-10pt}
\end{subfigure}
\hfill
\begin{subfigure}{0.33\textwidth}
    \includegraphics[width=\textwidth]{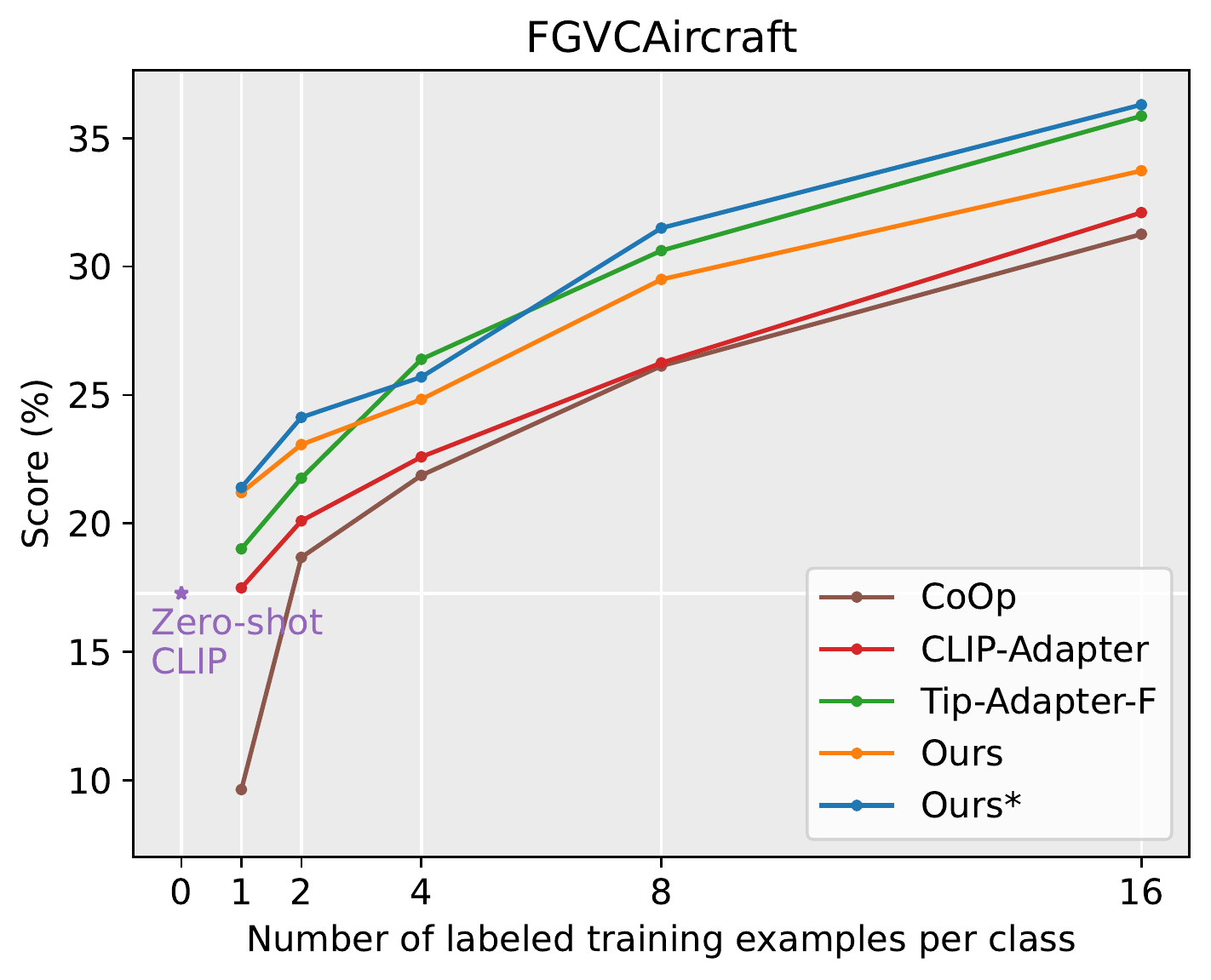}
    \vspace{-10pt}
\end{subfigure}
\hfill
\begin{subfigure}{0.33\textwidth}
    \includegraphics[width=\textwidth]{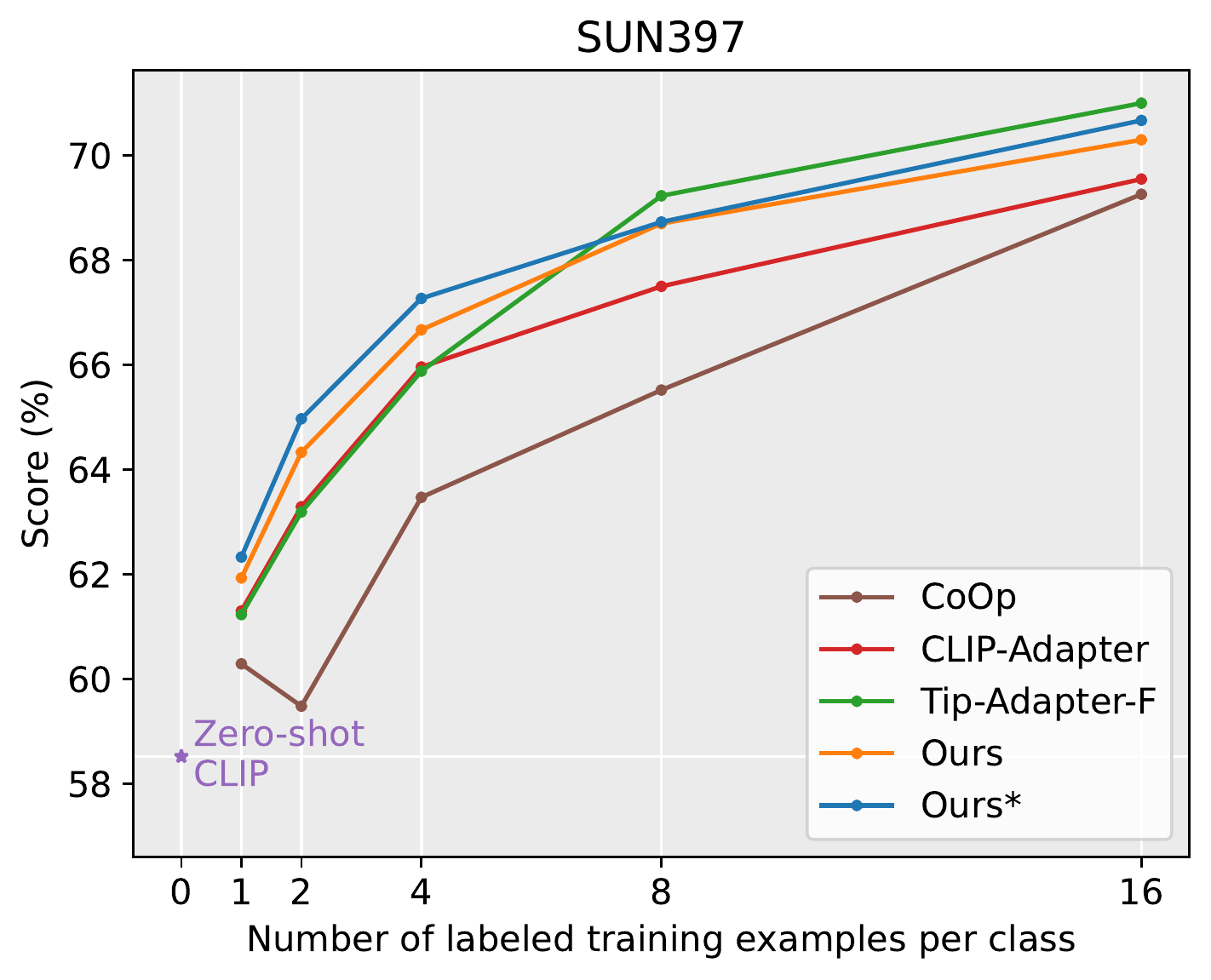}
    \vspace{-10pt}
\end{subfigure}

\begin{subfigure}{0.33\textwidth}
    \includegraphics[width=\textwidth]{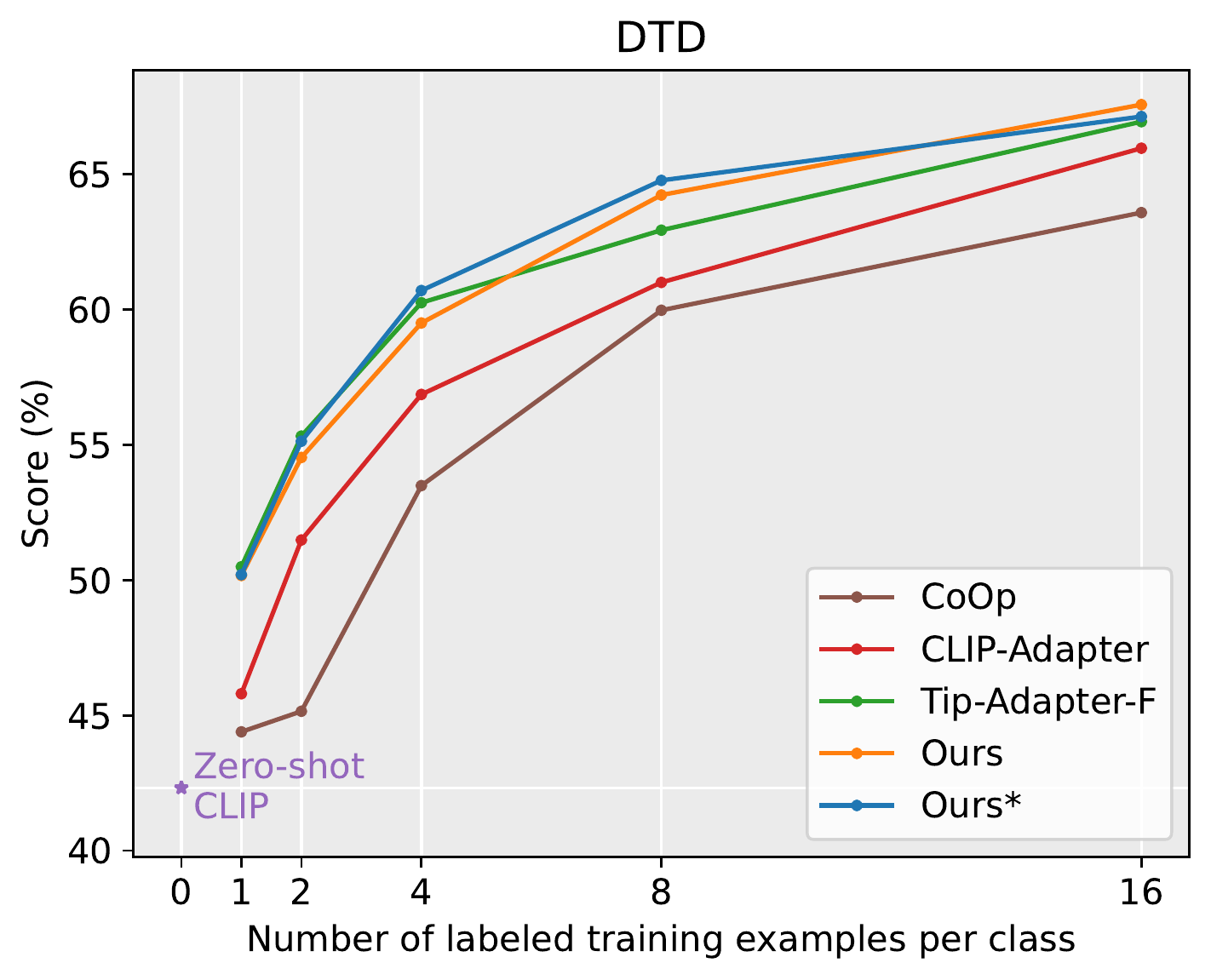}
\end{subfigure}
\hfill
\begin{subfigure}{0.33\textwidth}
    \includegraphics[width=\textwidth]{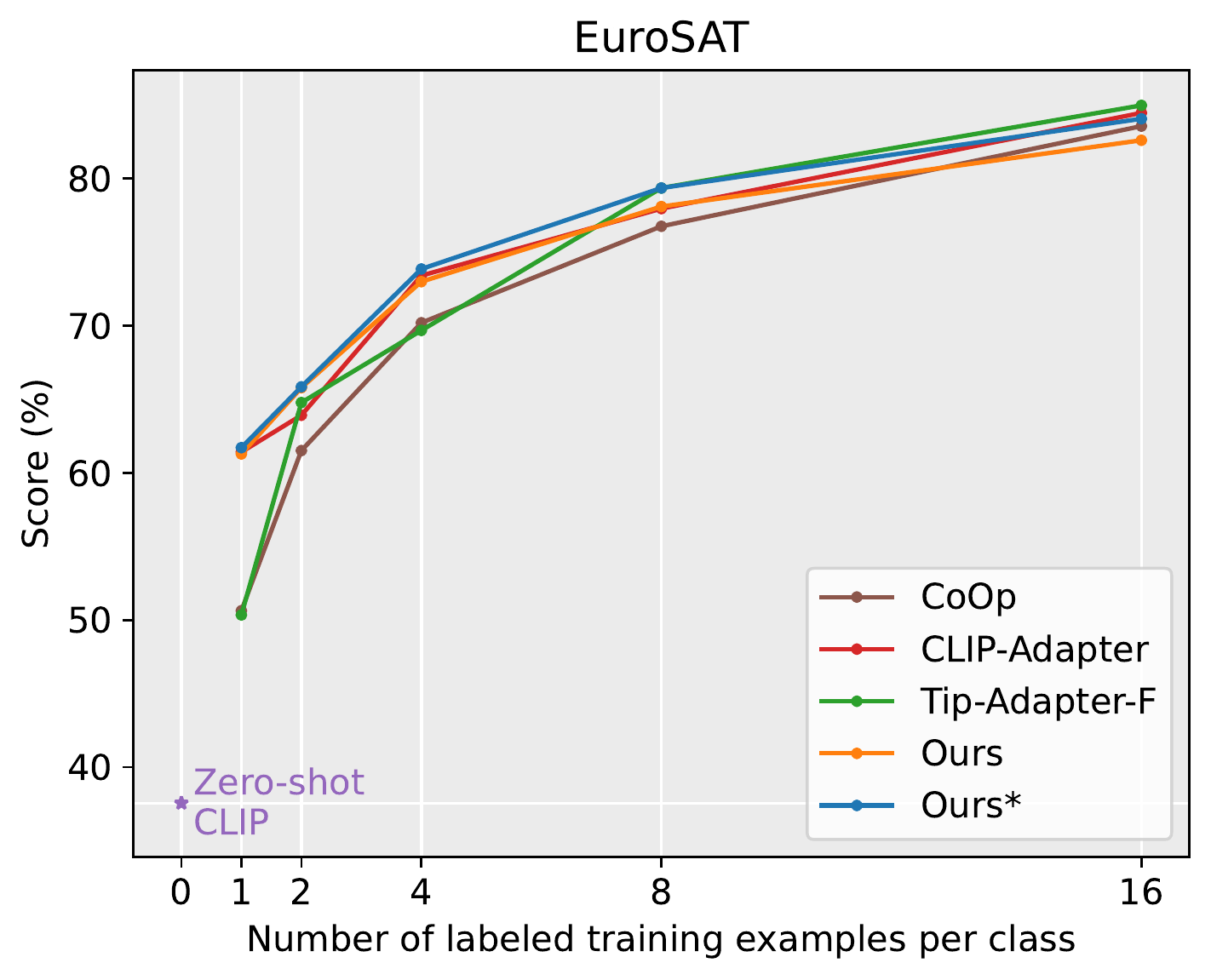}
\end{subfigure}
\hfill
\begin{subfigure}{0.33\textwidth}
    \includegraphics[width=\textwidth]{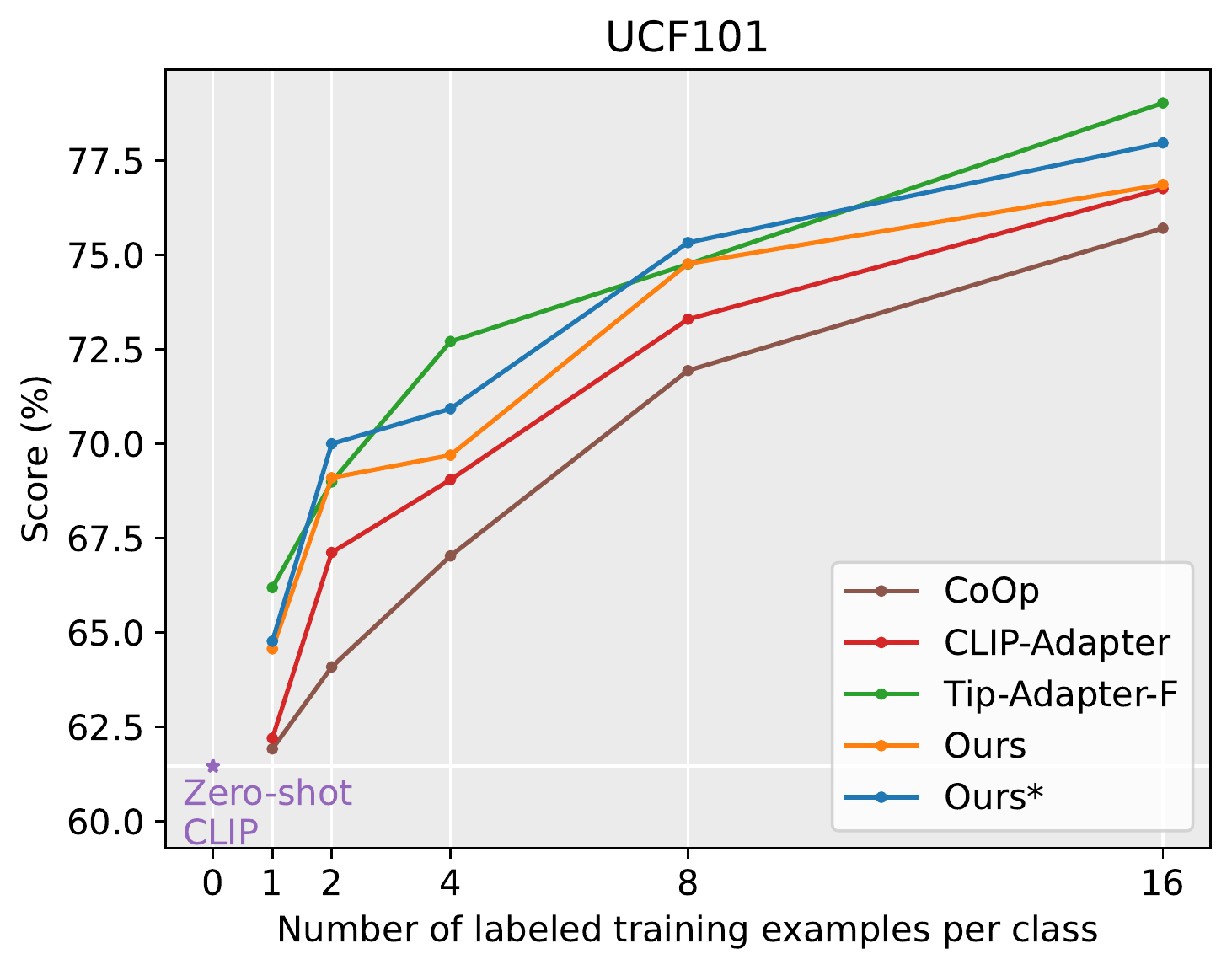}
\end{subfigure}
\end{adjustbox}
\caption{Performance comparison on few-shot learning, \emph{i.e.}, 1-/2-/4-/8-/16-shot, on 11 benchmark datasets. The top-left is the averaged accuracy over the 11 datasets. The full numerical results can be found in the supplementary.}
\label{fig:compare_11_datasets}
\vspace{-10pt}
\end{figure*}

\begin{table*}[ht]
    \centering
    \begin{adjustbox}{max width = 0.62\textwidth}
    \begin{tabular}{lcccccccc}
        \toprule
        \multirow{2}{*}{Method} & \multirow{2}{*}{Visual Backbone} & Source & \multicolumn{5}{c}{Target} \\
        \cmidrule(lr){3-3} \cmidrule(lr){4-8}
        & & ImageNet & -V2 & -Sketch & -A & -R & Average\\
        \midrule
        Zero-Shot CLIP \cite{radford2021learning} & \multirow{5}{*}{ResNet-50} & 58.18 & 51.34 & 33.32 & 21.65 & 56.00 & 40.58 \\
        Linear Probe CLIP \cite{radford2021learning} & & 55.87 & 45.97 & 19.07 & 12.74 & 34.86 & 28.16\\
        CoOp \cite{zhou2022learning} & & 62.95 & 55.11 & 32.74 & 22.12 & 54.96 & 41.23\\
        \rowcolor{Gray}
        Ours & & 64.75 & 56.47 & \textbf{35.83} & \textbf{22.80} & \textbf{60.70} & \textbf{43.95}\\
        \rowcolor{Gray}
        Ours* & & \textbf{65.73} & \textbf{57.00} & 34.43 & 21.50 & 58.13 & 42.77\\
        \midrule
        Zero-Shot CLIP \cite{radford2021learning} & \multirow{5}{*}{ResNet-101} & 61.62 & 54.81 & 38.71 & 28.05 & 64.38 & 46.49 \\
        Linear Probe CLIP \cite{radford2021learning} & & 59.75 & 50.05 & 26.80 & 19.44 & 47.19 & 35.87\\
        CoOp \cite{zhou2022learning} & & 66.60 & 58.66 & 39.08 & 28.89 & 63.00 & 47.41 \\
        \rowcolor{Gray}
        Ours & & 67.70 & 59.50 & \textbf{41.70} & \textbf{29.87} & \textbf{68.07} & \textbf{49.79}\\
        \rowcolor{Gray}
        Ours* & & \textbf{68.73} & \textbf{60.00} & 40.3 & 28.00 & 64.80 & 48.28\\
        \midrule
        Zero-Shot CLIP \cite{radford2021learning} & \multirow{5}{*}{ViT-B/32} & 62.05 & 54.79 & 40.82 & 29.57 & 65.99 & 47.79\\
        Linear Probe CLIP \cite{radford2021learning} & & 59.58 & 49.73 & 28.06 & 19.67 & 47.20 & 36.17\\
        CoOp \cite{zhou2022learning} & & 66.85 & 58.08 & 40.44 & 30.62 & 64.45 & 48.40\\
        \rowcolor{Gray}
        Ours & & 68.20 & 59.20 & \textbf{42.50} & \textbf{31.43} & \textbf{69.33} & \textbf{50.62}\\
        \rowcolor{Gray}
        Ours* & & \textbf{69.17} & \textbf{59.47} & 40.87 & 29.70 & 66.27 & 49.08\\
        \midrule
        Zero-Shot CLIP \cite{radford2021learning} & \multirow{5}{*}{ViT-B/16} & 66.73 & 60.83 & 46.15 & 47.77 & 73.96 & 57.18 \\
        Linear Probe CLIP \cite{radford2021learning} & & 65.85 & 56.26 & 34.77 & 35.68 & 58.43 & 46.29\\
        CoOp \cite{zhou2022learning} & & 71.92 & 64.18 & 46.71 & 48.41 & 74.32 & 58.41 \\
        \rowcolor{Gray}
        Ours & & 73.07 & 65.30 & \textbf{49.13} & \textbf{50.37} & \textbf{77.70} & \textbf{60.63}\\
        \rowcolor{Gray}
        Ours* & & \textbf{73.90} & \textbf{65.85} & 47.70 & 49.17 & 75.23 & 59.49\\
        \bottomrule
    \end{tabular}
    \end{adjustbox}
    \caption{Performance comparison on generalization (from ImageNet to ImageNet-V2/-Sketch/-A/-R) with multiple CLIP visual backbones.
    }
    \label{tab:DG}
    \vspace{-8pt}
\end{table*}
\section{Experiment}
\subsection{Setup}
We follow prior efficient transfer learning (ETL) work \cite{zhou2022learning,gao2021clip,zhang2022tip} to conduct a few-shot evaluation for ETL models on 11 benchmark datasets \ieno, ImageNet \cite{deng2009imagenet}, Caltech101 \cite{fei2004learning}, OxfordPets \cite{parkhi2012cats}, StanfordCars \cite{krause20133d}, Flowers102 \cite{nilsback2008automated}, Food101 \cite{bossard2014food}, FGVCAircraft \cite{maji2013fine}, SUN397 \cite{xiao2010sun}, DTD \cite{cimpoi2014describing}, EuroSAT \cite{helber2019eurosat} and UCF101 \cite{soomro2012ucf101}. Those datasets include a wide range of visual recognition tasks, \egno, classifying general objects, fine-grained objects, actions, scenes, \etcno.
Specifically, ETL models are trained using 1/2/4/8/16 shots per class from the training sets, respectively, and are tested on the full test sets. Additionally, following CoOp \cite{zhou2022learning}, we test the generalization performance of our models from ImageNet to its variants (ImageNetV2 \cite{recht2019imagenet}, ImageNet-Sketch \cite{wang2019learning}, ImageNet-A \cite{hendrycks2021natural} and ImageNet-R \cite{hendrycks2021many}).

\subsection{Implementation}
Our approach has two main components, \ieno, base classifier and task residual. For the base classifier, we develop two versions: \textit{regular base classifier} that directly adopts the weights of the text embeddings of the pre-trained CLIP (using pre-defined text templates following prior work \cite{radford2021learning,zhang2022tip}), and \textit{enhanced base classifier} obtained by tuning the text projection layer of CLIP on the target task before starting our task residual tuning\footnote{The enhanced base classifier is introduced to help us investigate whether there are consistent performance improvements when the base classifier is enhanced.}. Task residual is a matrix filled with learnable parameters that are initialized by zeros. Task residual has the same shape as the base classifier. We scale the task residual and element-wisely add it to the base classifier, as in Eq. \ref{eq:target cls}.
By default, the scaling factor $\alpha$ is set to 0.5 for all datasets except for Flowers102 using 1. Additionally, we explore the use of a learnable $\alpha$ later in our ablation study.
All models in this work are built upon the pre-trained CLIP models. Unless otherwise specified, we use the ResNet-50 \cite{he2016deep} version of CLIP which has a ResNet-50 backbone in the image branch. We train our models for 100 epochs for 1-/2-/4-shot experiments and 200 epochs for 8-/16-shot experiments, with batch size 256. Besides, the aforementioned enhanced base classifier is tuned for 50 epochs. Our models are optimized by Adam \cite{kingma2014adam} with an initial learning rate of 2$e$-3 except for ImageNet using 2$e$-4. Following CoOp \cite{zhou2022learning}, our optimization process adopts a cosine learning rate decay schedule and a warmup scheme (\ieno, fixing the learning rate in the first epoch to 1$e$-5). We test the performance of models after training, and all experimental results are averaged over three random seeds.

\subsection{Performance Comparison}
\subsubsection{Few-Shot Learning}
We develop two versions of our \ours, \ieno, \textit{Ours}/\textit{Ours*} using the regular/enhanced base classifier, respectively.
We compare our models with Zero-shot CLIP \cite{radford2021learning}, together with the state-of-the-art (SOTA) ETL methods including CoOp \cite{zhou2022learning}, CLIP-Adapter \cite{gao2021clip} and Tip-Adapter-F \cite{zhang2022tip}\footnote{Original Tip-Adapter \cite{zhang2022tip} has two versions: training-free Tip-Adapter and a fine-tuning version called Tip-Adapter-F. We report the one with higher performance, \ieno, Tip-Adapter-F.}. The comparison results of the 11 benchmark datasets are shown in Figure \ref{fig:compare_11_datasets}. Overall, both Ours and Ours* achieve the SOTA/SOTA-comparable performance on the 11 datasets across all few-shot settings and significantly surpass Zero-shot CLIP \cite{radford2021learning}. 
Besides, a few observations can be made.
First, with more shots, Tip-Adapter-F reaches a closer performance to ours, but as shown in Figure \ref{fig:params_acc} the number of its tunable parameters increases linearly with that of shots, substantially limiting its scalability. Second, our method shows inferior performance on OxfordPets and Food101. Training with a few shots on these two fine-grained datasets is prone to be overfitting, which is also found in CoOp \cite{zhou2022learning}. Finally, we found the proposed method is superior when needed to recognize a large amount of classes, \egno, 1000 classes of ImageNet and 397 classes of SUN397 in 1-shot case. This is because our TaskRes, not excessively biased to the pre-trained features, can better explore the task-specific knowledge even in such extreme scenarios.


\vspace{-10pt}
\subsubsection{Domain Generalization}
As pointed out by CoOp \cite{zhou2022learning}, ETL models that are trained on a specific domain are at risk of learning spurious correlations when generalizing to unseen domains. We evaluate our models concerning domain generalization. We train our models on 16-shot ImageNet and test the generalization performance of the trained models on four unseen ImageNet variant datasets (ImageNet-V2, -Sketch, -A and -R). As shown in Table \ref{tab:DG}, our models consistently outperform the compared models (Zero-shot CLIP \cite{radford2021learning}, Linear Probe CLIP \cite{radford2021learning} and CoOp \cite{zhou2022learning}) across various CLIP visual backbones (ResNet-50 \cite{he2016deep}, ResNet-101 \cite{he2016deep}, ViT-B/32 \cite{dosovitskiy2020image} and ViT-B/16 \cite{dosovitskiy2020image}).
We also observe that our TaskRes using the enhanced base classifier (Ours*) achieves higher accuracy than that using the regular base classifier (Ours) on the source dataset while reducing a certain degree of generalization\tcr{, which is due to a slight overfitting on the source dataset.}

\subsection{Ablation Study}
\paragraph{Importance of prior knowledge preservation and prior-independent tuning.}
We build four target classifiers $\mathbf{t}'$ to investigate the importance of prior knowledge preservation and prior-independent tuning (where the tunable parameters are not \tcb{dependent} on the pre-trained features), \ieno, original pre-trained classifier $\mathbf{t}'=\mathbf{t}$ (\ieno, our regular base classifier), directly-adapted classifier $\mathbf{t}'=\phi_{\omega}(\mathbf{t})$, adapter-style classifier $\mathbf{t}'=\mathbf{t}+\alpha\phi_{\omega}(\mathbf{t})$ and our TaskRes classifier $\mathbf{t}'=\mathbf{t}+\alpha\mathbf{x}$. We conduct experiments on ImageNet across all few-shot settings, using the same scaling factor $\alpha=0.5$.
The results in Table \ref{tab:composition} support our motivations that (i) explicit preservation of prior knowledge matters as the directly-adapted classifier performs much worse than the original pre-trained classifier and adapter-style classifier, and (ii) tuning prior-independent parameters is more effective than the dependent one as the TaskRes classifier obviously surpasses the adapter-style classifier.

\begin{table}[h]
\centering
\begin{adjustbox}{max width=0.40\textwidth}
    \begin{tabular}{lccccc}
        \toprule
             Target Classifier $\mathbf{t}'$ & 1-shot & 2-shot & 4-shot & 8-shot & 16-shot  \\
             \midrule
             $\mathbf{t}$ & 60.33 & 60.33 & 60.33 & 60.33 & 60.33\\
             $\phi_{\omega}(\mathbf{t})$ & 28.03 & 34.90 & 42.57 & 48.17 & 54.13\\
             $\mathbf{t} + \alpha\phi_{\omega}(\mathbf{t})$ & 61.03 & 61.23 & 61.27 & 62.00 & 63.17 \\
             $\mathbf{t} + \alpha\mathbf{x}$ & \textbf{61.43} & \textbf{62.17} & \textbf{62.93} & \textbf{64.03} & \textbf{64.75} \\
        \bottomrule
    \end{tabular}
\end{adjustbox}
    \caption{Ablation study of various combinations of three main components: base classifier $\mathbf{t}$, adapter module $\phi_{\omega}(\cdot)$ and our ``task residual'' $\mathbf{x}$, for building a target classifier $\mathbf{t}'$. The experiment is conducted on ImageNet.}
    \label{tab:composition}
    \vspace{-15pt}
\end{table}

\paragraph{Effectiveness of task residual learning.}
We conduct an ablation study of the effect of introducing our task residual learning (TaskRes). Table \ref{tab:effect} shows the few-shot results averaged over 11 datasets of the regular base classifier and that equipped with our TaskRes (Ours), together with the enhanced counterparts. Our TaskRes notably improves the regular base, \egno, with 16.14\% accuracy gain in 16-shot case. Similarly, the performance gain is also impressive even on the enhanced base. This indicates that TaskRes can always benefit a pre-trained base when adapted to downstream tasks.

\begin{table}[h]
    \centering
    \begin{adjustbox}{max width=0.45\textwidth}
    \begin{tabular}{lccccc}
        \toprule
         Classifier & 1-shot & 2-shot & 4-shot & 8-shot & 16-shot \\
         \midrule
         Regular Base & 58.96 & 58.96 & 58.96 & 58.96 & 58.96 \\
         Regular Base + TaskRes & \textbf{64.04} & \textbf{67.02} & \textbf{69.61} & \textbf{72.76} &  \textbf{75.10} \\
         \midrule
         Enhanced Base & 61.11 & 62.82 & 64.44 & 69.91 & 73.09 \\
         Enhanced Base + TaskRes & \textbf{64.28} & \textbf{67.55} & \textbf{70.28} & \textbf{73.35} & \textbf{75.78} \\
         \bottomrule
    \end{tabular}
    \end{adjustbox}
    \caption{Exploring the effectiveness of the proposed task residual tuning (TaskRes). The accuracy results averaged on 11 datasets are reported under all shots. Refer to Appendix \ref{apx_ablation} for the full comparison results of the 11 datasets.}
    \label{tab:effect}
    \vspace{-15pt}
\end{table}

\paragraph{Variants of TaskRes.}
The proposed ``task residual'' can also be applied to the image branch of CLIP\tcb{, denoted as \textit{\ours-I}. We also denote the default \ours~applied on the text branch as \textit{\ours-T} (\ieno, Ours) and on the both branches as \textit{\ours-I$\&$T}. We compare their performance on ImageNet in Table \ref{tab:variants}.}
\ours-T consistently outperforms \ours-I, as shown in Table \ref{tab:variants}. The performance of \ours-I increases much slower than \ours-T as the number of shots increases. This is because \ours-I tends to encounter overfitting where training and testing image embeddings are different. Tuning on text embeddings can avoid the issue. Additionally, the diversity of image embeddings is larger than that of text embeddings, making it difficult to learn task-level information.
\tcb{\ours-I$\&$T performs slightly worse than \ours-T. This might be caused by a misalignment between the updated image embeddings and our base classifier.}

\begin{table}[h]
    \centering
    \begin{adjustbox}{max width=0.38\textwidth}
    \begin{tabular}{lccccc}
        \toprule
         Model & 1-shot & 2-shot & 4-shot & 8-shot & 16-shot \\
         \midrule
         Regular Base & 60.33 & 60.33 & 60.33 & 60.33 & 60.33 \\
         TaskRes-I & 61.20 & 61.27 & 61.47 & 61.53 & 61.60 \\
         TaskRes-T & 61.43 & \textbf{62.17} & \textbf{62.93} & \textbf{64.03} &  \textbf{64.75} \\
         \tcb{TaskRes-I$\&$T} & \textbf{61.53} & 62.10 & 62.73 & 63.83 & 64.67 \\
         \bottomrule
    \end{tabular}
    \end{adjustbox}
    \caption{\tcb{Performance comparison of three TaskRes variants on ImageNet: TaskRes applied on CLIP's image branch (TaskRes-I), text branch (TaskRes-T) and both branches (TaskRes-I$\&$T).}}
    \label{tab:variants}
    \vspace{-25pt}
\end{table}

\paragraph{Scaling factor.} 
\tcb{We investigate the effect of scaling factor $\alpha$ and show the results of both manually assigned values and a learned value in Table \ref{tab:alpha}. For the manually set $\alpha$, we observe} a notable improvement with TaskRes scaled even by 0.1, while the highest accuracy is obtained with $\alpha=0.5$. 
\tcr{In contrast, the learnable $\alpha$ can further boost the performance. We show more analyses and results in Appendix \ref{apx_ablation}.}

\begin{table}[h]
    \centering
    \begin{adjustbox}{max width=0.42\textwidth}
    \begin{tabular}{lccccccc}
        \toprule
         $\alpha$ & 0 & 0.1 & 0.3 & 0.5 & 0.7 & 1.0 & Learned\\
         \midrule
         Acc. & 58.96 & 61.86 & 63.76 & \textbf{63.97} & 63.67 & 63.02 & \textbf{64.01} \\
        \bottomrule
    \end{tabular}
    \end{adjustbox}
    \caption{Ablation study of scaling factor $\alpha$. The results are \tcb{averaged over} 11 datasets in 1-shot case.}
    \label{tab:alpha}
    \vspace{-10pt}
\end{table}

\paragraph{Visual backbone.}
We further evaluate models on various CLIP visual backbones, namely, ResNet-50 (RN50), ResNet-101 (RN101), ViT-B/32 and ViT-B/16, as shown in Table \ref{tab:backbone}. Our method shows consistent superiority against other alternatives regardless of which backbone is used.

\begin{table}[h]
    \centering
    \begin{adjustbox}{max width=0.40\textwidth}
    \begin{tabular}{lcccc}
        \toprule
        Method & RN50 & RN101 & ViT-B/32 & ViT-B/16 \\
        \midrule
        Zero-shot CLIP \cite{radford2021learning} & 58.18 & 61.62 & 62.05 & 66.73 \\
        CoOp \cite{zhou2022learning} & 62.95 & 66.60 & 66.85 & 71.92 \\
        CLIP-Adapter \cite{gao2021clip} & 63.59 & 65.39 & 66.19 & 71.13 \\
        Tip-Adapter-F \cite{zhang2022tip} & 65.44 & 68.56 & 68.65 & 73.69 \\
        \rowcolor{Gray}
        Ours & 64.75 & 67.70 & 68.20 & 73.07 \\
        \rowcolor{Gray}
        Ours* & \textbf{65.73} & \textbf{68.73} & \textbf{69.17} & \textbf{73.90}\\
        \bottomrule
    \end{tabular}
    \end{adjustbox}
    \caption{Results of  CLIP visual backbones on 16-shot ImageNet.}
    \label{tab:backbone}
    \vspace{-2pt}
\end{table}

\vspace{-8pt}
\subsection{Investigation of Learned Task Residual}
\begin{figure}[t]
	\begin{center}
		\includegraphics[scale=0.44]{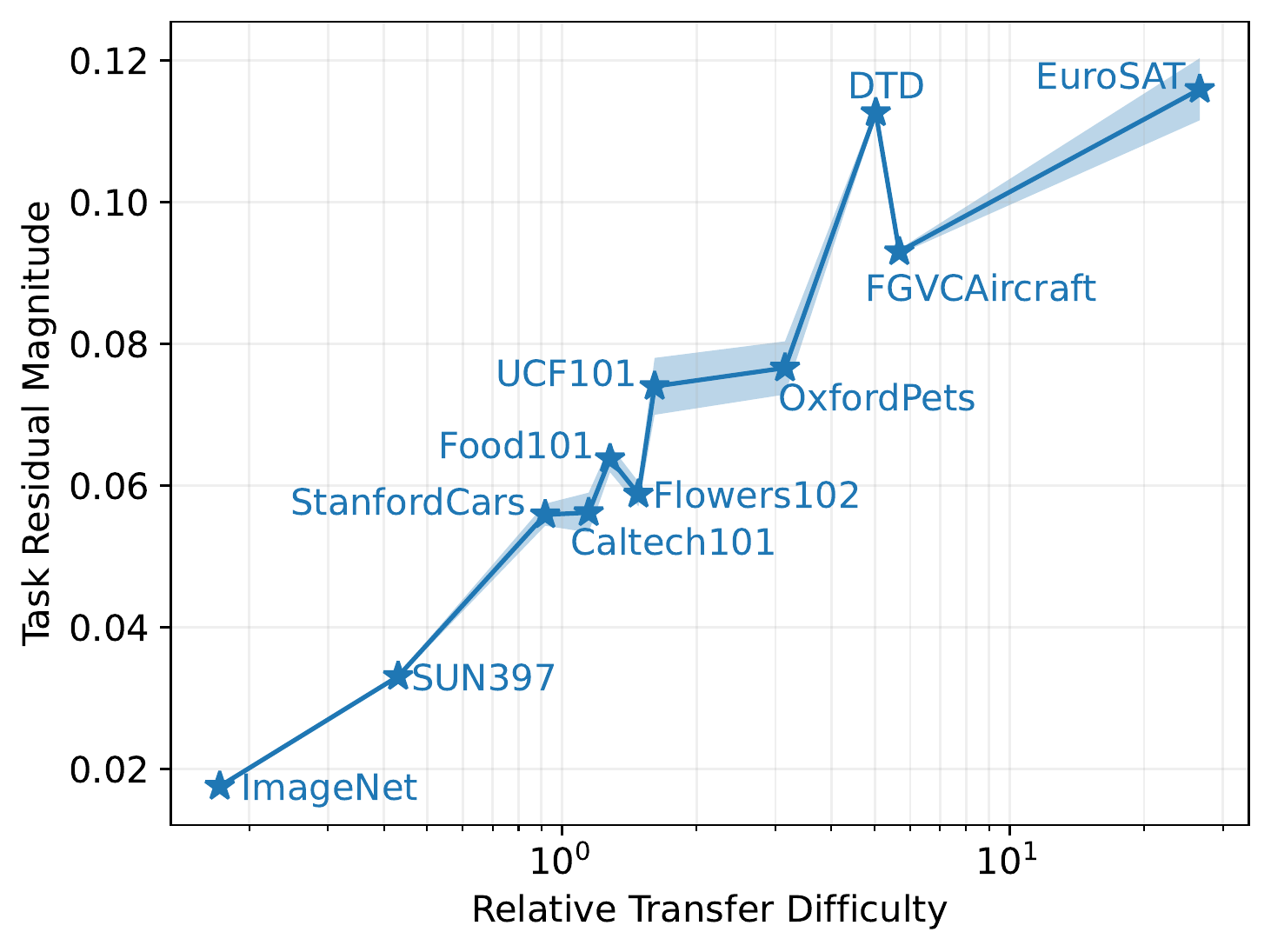} 
	\end{center}
        \vspace{-18pt}
	\caption{The magnitude of learned task residual increases with relative transfer difficulty (in logarithmic scale). The shadow indicates the standard deviation regarding random seeds.}
	\label{fig:taskres_diff}
        \vspace{-12pt}
\end{figure}

We investigate whether the learned task residuals are in fact related to the difficulty of transferring CLIP to downstream tasks. We first define \textit{relative transfer difficulty} to assess the difficulty of transferring a pre-trained model to a target task. Concretely, the relative transfer difficulty is defined as the ratio of baseline classifier precision to the zero-shot precision of the pre-trained model. We use a random classifier to serve as the baseline classifier where the precision of it equals to $1/K$ ($K$ is the number of classes of the downstream task). We then can calculate the relative transfer difficulty of CLIP regarding the 11 benchmark datasets based on the random classifier and Zero-shot CLIP. We count the average magnitudes of the learned task residual of each dataset in 16-shot setting and find that they are highly related to relative transfer difficulty, as shown in Figure \ref{fig:taskres_diff}. This verifies the reliability of the proposed task residual. There are some interesting findings that may be counter-intuitive, \egno,  transferring CLIP to ImageNet (with 1000 classes) is the simplest and to EuroSAT (with 10 classes) is the most difficult. This is because the inherent difficulty of ImageNet is large but the data distribution and containing objects are ``familiar'' to CLIP, whereas EuroSAT is the opposite. Besides, there is more space for \ours~to play its role in the tasks with large relative transfer difficulty, \egno, EuroSAT (1-shot accuracy gain: 23.71\%) and DTD (1-shot accuracy gain: 7.85\%). \tcb{More  results and analyses can be found in Appendix \ref{apx_vis}.}

\section{Conclusion, Limitation and Future Work}
In this work, we propose a new approach for tuning VLMs, \ieno, \ours. The proposed \ours~performs efficient transfer learning (ETL) on VLMs by explicitly decoupling the classifier into two key parts: undamaged base classifier with rich prior knowledge and task residual independent to the base classifier for better exploring task-specific knowledge. Interestingly, the magnitude of learned task residual is highly related to the difficulty of transferring pre-trained VLMs to a target downstream task. This may inspire the community to consider ETL from a new perspective, \egno, modeling ``task-to-task transfer difficulty'' for ETL. Extensive experiments have demonstrated the effectiveness of our \ours, despite its simplicity. 

Nevertheless, our method bears some limitations. For instance, our method encounters negative transfer on two datasets, \ieno, OxfordPets (1-shot) and Food101 (1-/2-/4-/8-shot). 
We conjecture that this happens with two conditions: (i) the downstream tasks are of high relative transfer difficulty as shown in Figure \ref{fig:taskres_diff}, and (ii) Zero-shot CLIP has already achieved a fair precision on them. \tcb{Besides, the assessment of transfer difficulty in this work is heuristic.}

With the rapid advancement of foundation models \cite{bommasani2021opportunities}, \tcb{it is becoming increasingly crucial to establish precise and reliable metrics for assessing the transfer difficulty of pre-trained foundation models to downstream tasks. A comprehensive study of transfer difficulty, which includes distribution analyses \cite{li2022elevater}, is in high demand.} 
\tcr{Moreover, we can extend the transfer difficulty to a concept-wise level, and investigate the correlation between performance and the occurrence frequency of visual concepts by exploring the CLIP models trained on specific datasets, \egno, SLIP \cite{mu2022slip} trained on YFCC15M \cite{radford2018improving, thomee2016yfcc100m}.} \tcb{We will explore these in future work.}

\section*{Acknowledgment}
This research is supported by the Ministry of Education, Singapore, under its Academic Research Fund Tier~2
(Award Number:~MOE-T2EP20122-0006),
the National Research Foundation, Singapore under its Medium Sized Center for Advanced Robotics Technology Innovation,
NSFC under Grant U1908209 and 62021001, and ZJNSFC under Grant LQ23F010008.

\newpage
{\small
\bibliographystyle{ieee_fullname}
\bibliography{egbib}
}

\clearpage

\begin{strip}
\begin{center}
\textbf{\Large Task Residual for Tuning Vision-Language Models \\
(Appendix)}
\end{center}
\end{strip}

\appendix

\section{Summary of Datasets}
In our main text, we evaluate the proposed method for few-shot learning tasks on 11 benchmark datasets and domain generalization tasks on ImageNet to its variants (ImageNet-V2, -Sketch, -A and -R). We summarize the dataset information in Table \ref{tab:dataset_summary}. 

Specifically, the datasets for the few-shot evaluation are composed of diverse genres, such as recognition of generic objects, fine-grained objects, scenes, textural images and satellite images.
The diversity can better verify the effectiveness and robustness of the proposed method.
To be consistent with previous works \cite{zhou2022learning, gao2021clip, zhang2022tip}, the “BACKGROUND Google” and “Faces easy” classes are removed in Caltech101  \cite{fei2004learning}. 
We also list the used templates \cite{zhang2022tip} for the text-based classifier construction based on the CLIP's \cite{radford2021learning} text branch.

For the generalization datasets, ImageNet-V2 and ImageNet-Sketch share the same label space with ImageNet (1000 classes), while the label spaces of ImageNet-A (200 classes) and ImageNet-R (200 classes) are both sub-spaces of the ImageNet label space.
The variants of ImageNet contain substantially different data distributions (See Table \ref{tab:dataset_summary} Description) from ImageNet, which makes them satisfactory domain generalization benchmarks.
Following CoOp \cite{zhou2022learning}, we choose ImageNet as the source domain data while the variants as the target one.

\begin{table*}[h]
    \centering
    \begin{adjustbox}{max width = 1.0\textwidth}
    \begin{tabular}{lccccc}
        \toprule
        Name & Number of Classes & Size (Train / Val / Test) & Description & Template\\
        \midrule
        ImageNet \cite{deng2009imagenet} & 1000 & 1.28M / - /50000 &  Recognition of generic objects & Ensemble of 7 selected templates \\
        Caltech101 \cite{fei2004learning} & 100 & 4128 / 1649 / 2465 & Recognition of generic objects & ``a photo of a [class].'' \\
        OxfordPets \cite{parkhi2012cats} & 37 & 2944 / 736 / 3669 & Fine-grained classification of pets & ``a photo of a [class], a type of pet.'' \\
        StanfordCars \cite{krause20133d} & 196 & 6509 / 1635 / 8041 & Fine-grained classification of cars & ``a photo of a [class].'' \\
        Flowers102 \cite{nilsback2008automated} & 102 & 4093 / 1633 / 2463 & Fine-grained classification of flowers & ``a photo of a [class], a type of flower.'' \\
        Food101 \cite{bossard2014food} & 101 & 50500 / 20200 / 30300 & Fine-grained classification of foods & ``a photo of a [class], a type of food.'' \\
        FGVCAircraft \cite{maji2013fine} & 100 & 3334 / 3333 / 3333 & Fine-grained classification of aircrafts &``a photo of a [class], a type of aircraft.'' \\
        SUN397 \cite{xiao2010sun} & 397 & 15880 / 3970 / 19850 & Scene classification & ``a photo of a [class].'' \\
        DTD \cite{cimpoi2014describing} & 47 & 2820 / 1128 / 1692 & Texture classification & ``[class] texture.'' \\
        EuroSAT \cite{helber2019eurosat} & 10 & 13500 / 5400 / 8100 & Land use \& cover classification with satellite images & ``a centered satellite photo of [class].'' \\
        UCF101 \cite{soomro2012ucf101} & 101 & 7639 / 1898 / 3783 & Action recognition & ``a photo of a person doing [class].'' \\
        \midrule
        ImageNet-V2 \cite{recht2019imagenet} & 1000 & - / - / 10000 & New test data for ImageNet & Ensemble of 7 selected templates\\
        ImageNet-Sketch \cite{wang2019learning} & 1000 & - / - / 50889 & Sketch-style images of ImageNet classes & Ensemble of 7 selected templates\\
        ImageNet-A \cite{hendrycks2021natural} & 200 & - / - / 7500 & Natural adversarial examples of 200 ImageNet classes & Ensemble of 7 selected templates\\
        ImageNet-R \cite{hendrycks2021many} & 200 & - / - / 30000 & Renditions of 200 ImageNet classes & Ensemble of 7 selected templates\\
        \bottomrule
    \end{tabular}
    \end{adjustbox}
    \caption{Summary of 11 datasets for few-shot learning and 4 target datasets of domain generalization. The 7 selected templates \cite{zhang2022tip} for ImageNet series datasets are ``itap of a [class].'', ``a bad photo of the [class].'', ``a origami [class].'', ``a photo of the large [class].'', ``a [class] in a video game.'', ``art of the [class].'' and ``a photo of the small [class].''.
    }
    \label{tab:dataset_summary}
\end{table*}

\section{More Experimental Results and Analyses}
\subsection{Few-Shot Learning}
The full numerical results of Figure \ref{fig:compare_11_datasets} in the main text are presented in Table \ref{tab:numerical_compare_11}.
Note that the results of Tip-Adapter-F \cite{zhang2022tip} are slightly different from their original paper. The original Tip-Adapter-F tests their models per epoch of training and chooses the best one to report the performance, while other methods such as CoOp test model until the training is done. 
To make the comparison fair, we re-run the official code of Tip-Adapter-F and test its models at the end of training.
Overall, our method achieves the best averaged performance across all shot settings and datasets. 
In particular, our method reaches the best performance on ImageNet, Caltech101 and StanfordCars for all shot settings and on Flowers102, FGVCAircraft, SUN397, DTD and EuroSAT for most shot settings. 
Additionally, despite having limited tunable parameters, the proposed method can always benefit from the expansion of training data, \ieno, from 1-shot to 16-shot with the averaged gains from 5.08\% to 16.14\%. In contrast, Tip-Adapter-F \cite{zhang2022tip} achieves similar performance by linearly increasing the tunable parameters with the number of shots.
\begin{table*}[h]
    \centering
    \begin{adjustbox}{max width = 1.0\textwidth}
    \begin{tabular}{lccccccccccccc}
        \toprule
        Method & Setting & ImageNet & Caltech101 & OxfordPets & StanfordCars & Flowers102 & Food101 & FGVCAircraft & SUN397 & DTD & EuroSAT & UCF101 & Average \\
        \midrule
        Zero-Shot CLIP \cite{radford2021learning} & \multirow{6}{*}{1-shot} & 58.18 & 86.29 & 85.77 & 55.61 & 66.14 & \textbf{77.31} & 17.28 & 58.52 & 42.32 & 37.56 & 61.46 & 58.77  \\
        CoOp \cite{zhou2022learning} & & 57.15 & 87.53 & 85.89 & 55.59 & 68.12 & 74.32 & 9.64 & 60.29 & 44.39 & 50.63 & 61.92 & 59.59 \\
        CLIP-Adapter \cite{gao2021clip} & & 61.20 & 88.60 & 85.99 & 55.13 & 73.49 & 76.82 & 17.49 & 61.30 & 45.80 & 61.40 & 62.20 & 62.67 \\
        Tip-Adapter-F \cite{zhang2022tip} & & 60.88 & \textbf{88.80} & \textbf{86.04} & 56.78 & \textbf{81.17} & 76.22 & 19.01 & 61.23 & \textbf{50.49} & 50.34 & \textbf{66.19} & 63.38 \\
        \rowcolor{Gray}
        Ours & & 61.43 & \textbf{88.80} & 83.50 & 58.77 & 78.77 & 74.03 & 21.20 & 61.93 & 50.17 & 61.27 & 64.57 & 64.04 \\
        \rowcolor{Gray}
        Ours* & & \textbf{61.90} & \textbf{88.80} & 83.60 & \textbf{59.13} & 79.17 & 74.03 & \textbf{21.40} & \textbf{62.33} & 50.20 & \textbf{61.70} & 64.77 & \textbf{64.28} \\
        \midrule
        Zero-Shot CLIP \cite{radford2021learning} & \multirow{6}{*}{2-shot} & 58.18 & 86.29 & 85.77 & 55.61 & 66.14 & \textbf{77.31} & 17.28 & 58.52 & 42.32 & 37.56 & 61.46 & 58.77  \\
        CoOp \cite{zhou2022learning} & & 57.81 & 87.93 & 82.64 & 58.28 & 77.51 & 72.49 & 18.68 & 59.48 & 45.15 & 61.50 & 64.09 & 62.32 \\
        CLIP-Adapter \cite{gao2021clip} & & 61.52 & 89.37 & \textbf{86.73} & 58.74 & 81.61 & 77.22 & 20.10 & 63.29 & 51.48 & 63.90 & 67.12 & 65.55 \\
        Tip-Adapter-F \cite{zhang2022tip} & & 61.57 & 89.61 & 86.06 & 61.13 & 85.40 & 77.05 & 21.76 & 63.19 & \textbf{55.32} & 64.76 & 68.99 & 66.80 \\
        \rowcolor{Gray}
        Ours & & 62.17 & 90.13 & 84.43 &	62.77 &	85.63 & 75.30 &	23.07 &	64.33 &	54.53 &	65.77 &	69.10 &	67.02 \\
        \rowcolor{Gray}
        Ours* & & \textbf{62.63} & \textbf{90.27} & 84.63 & \textbf{63.70} & \textbf{86.57} & 75.17 & \textbf{24.13} & \textbf{64.97} & 55.13 & \textbf{65.83} & \textbf{70.00} & \textbf{67.55} \\
        \midrule
        Zero-Shot CLIP \cite{radford2021learning} & \multirow{6}{*}{4-shot} & 58.18 & 86.29 & 85.77 & 55.61 & 66.14 & 77.31 & 17.28 & 58.52 & 42.32 & 37.56 & 61.46 & 58.77 \\
        CoOp \cite{zhou2022learning} & & 59.99 & 89.55 & 86.70 & 62.62 & 86.20 & 73.33 & 21.87 & 63.47 & 53.49 & 70.18 & 67.03 & 66.77 \\
        CLIP-Adapter \cite{gao2021clip} & & 61.84 & 89.98 & \textbf{87.46} & 62.45 & 87.17 & \textbf{77.92} & 22.59 & 65.96 & 56.86 & 73.38 & 69.05 & 68.61 \\
        Tip-Adapter-F \cite{zhang2022tip} & & 62.62 & 90.87 & 86.46 & 64.86 & 89.53 & 77.46 & \textbf{26.39} & 65.88 & 60.25 & 69.66 & \textbf{72.71} & 69.70 \\
        \rowcolor{Gray}
        Ours & & 62.93 & 90.63 & 86.27 & 66.50 & 89.50 & 76.23 & 24.83 & 66.67 & 59.50 & 72.97 & 69.70 & 69.61 \\
        \rowcolor{Gray}
        Ours* & & \textbf{63.57} & \textbf{90.97} & 86.33 & \textbf{67.43} & \textbf{90.20} & 76.10 & 25.70 & \textbf{67.27} & \textbf{60.70} & \textbf{73.83} & 70.93 & \textbf{70.28} \\
        \midrule
        Zero-Shot CLIP \cite{radford2021learning} & \multirow{6}{*}{8-shot} & 58.18 & 86.29 & 85.77 & 55.61 & 66.14 & 77.31 & 17.28 & 58.52 & 42.32 & 37.56 & 61.46 & 58.77 \\
        CoOp \cite{zhou2022learning} & & 61.56 & 90.21 & 85.32 & 68.43 & 91.18 & 71.82 & 26.13 & 65.52 & 59.97 & 76.73 & 71.94 & 69.89 \\
        CLIP-Adapter \cite{gao2021clip} & & 62.68 & 91.40 & 87.65 & 67.89 & 91.72 & \textbf{78.04} & 26.25 & 67.50 & 61.00 & 77.93 & 73.30 & 71.40 \\
        Tip-Adapter-F \cite{zhang2022tip} & & 64.15 & 91.70 & \textbf{88.28} & 69.51 & 91.00 & 77.90 & 30.62 & \textbf{69.23} & 62.93 & \textbf{79.33} & 74.76 & 72.67 \\
        \rowcolor{Gray}
        Ours & & 64.03 & 92.23 & 87.07 & 70.57 & 94.30 & 76.90 & 29.50 & 68.70 & 64.23 & 78.07 & 74.77 & 72.76 \\
        \rowcolor{Gray}
        Ours* & & \textbf{64.67} & \textbf{92.40} & 87.17 & \textbf{71.83} & \textbf{94.73} & 76.40 & \textbf{31.50} & 68.73 & \textbf{64.77} & \textbf{79.33} & \textbf{75.33} & \textbf{73.35} \\
        \midrule
        Zero-Shot CLIP \cite{radford2021learning} & \multirow{6}{*}{16-shot} &  58.18 & 86.29 & 85.77 & 55.61 & 66.14 & 77.31 & 17.28 & 58.52 & 42.32 & 37.56 & 61.46 & 58.77 \\
        CoOp \cite{zhou2022learning} & & 62.95 & 91.83 & 87.01 & 73.36 & 94.51 & 74.67 & 31.26 & 69.26 & 63.58 & 83.53 & 75.71 & 73.42 \\
        CLIP-Adapter \cite{gao2021clip} & & 63.59 & 92.49 & 87.84 & 74.01 & 93.90 & \textbf{78.25} & 32.10 & 69.55 & 65.96 & 84.43 & 76.76 & 74.44 \\
        Tip-Adapter-F \cite{zhang2022tip} & & 65.44 & 92.63 & \textbf{88.18} & 75.75 & 94.23 & 78.11 & 35.86 & \textbf{71.00} & 66.94 & \textbf{84.94} & \textbf{79.03} & 75.65 \\
        \rowcolor{Gray}
        Ours & & 64.75 & 92.90 & 88.10 & 74.93 & \textbf{96.10} & 78.23 & 33.73 & 70.30 & \textbf{67.57} & 82.57 & 76.87 & 75.10 \\
        \rowcolor{Gray}
        Ours* & & \textbf{65.73} & \textbf{93.43} & 87.83 & \textbf{76.83} & 96.03 & 77.60 & \textbf{36.30} & 70.67 & 67.13 & 84.03 & 77.97 & \textbf{75.78} \\
        \bottomrule
    \end{tabular}
    \end{adjustbox}
    \caption{Full numerical results of performance comparison on few-shot learning.
    }
    \label{tab:numerical_compare_11}
\end{table*}

\subsection{Training Efficiency}
\tcb{
Our TaskRes is not only parameter- and data- efficient but also highly efficient in training. As illustrated in Figure \ref{fig:motivation} in the main text, the high training efficiency is attributed to the absence of additional network modules (as in adapter-style tuning \cite{gao2021clip}) and the elimination of the need to run the text encoder every time (as in prompt tuning \cite{zhou2022learning}).} 
\tcr{In particular, the quantitative results show that TaskRes needs merely 11 minutes, much less than 121 minutes used in prompt tuning and 16 minutes in adapter-style tuning, when training models on 4-shot ImageNet with a single GeForce RTX 3090 GPU.
}

\subsection{Ablation Study}\label{apx_ablation}
\paragraph{Ablation study of \ours~effectiveness.} We present the full results of the ablation study of our \ours~effectiveness across 11 benchmark datasets in Table \ref{tab:taskres_effect}. Our \ours~achieves notable improvements over both the regular and enhanced base classifiers across almost all the datasets. 
When equipping the regular base classifier with our proposed \ours, the accuracy of the model is improved by 5.08\%, 8.06\%, 10.65\%, 13.80\% and 16.14\% for 1-, 2-, 4-, 8- and 16-shot settings, respectively. 
For the model based on the enhanced base classifier, our method still brings accuracy gains of 3.17\%, 4.73\%, 5.84\%, 3.44\% and 2.69\% for the above settings, respectively.
However, as mentioned in the limitations (in main text), we observe a negative transfer on OxfordPets and Food101, similar to CoOp \cite{zhou2022learning}. 
This negative transfer gap decreases with the number of shots increasing, which suggests that for these two datasets, learning the task-specific information is more difficult than other datasets, so more shots are needed.

\begin{table*}[h]
    \centering
    \begin{adjustbox}{max width = 1.0\textwidth}
    \begin{tabular}{llcccccccccccc}
        \toprule
        Setting & Method & ImageNet & Caltech101 & OxfordPets & StanfordCars & Flowers102 & Food101 & FGVCAircraft & SUN397 & DTD & EuroSAT & UCF101 & Average \\
        \midrule
        \multirow{4}{*}{1-shot} & Regular Base & 60.33 & 86.29 & \textbf{85.77} & 55.61 & 66.14 & \textbf{77.31} & 17.28 & 58.52 & 42.32 & 37.56 & 61.46 & 58.96   \\
        & Regular Base + \ours & \textbf{61.43} & \textbf{88.80} & 83.50 & \textbf{58.77} & \textbf{78.77} & 74.03 & \textbf{21.20} & \textbf{61.93} & \textbf{50.17} & \textbf{61.27} & \textbf{64.57} & \textbf{64.04}   \\
        \cmidrule(r){2-14}
        & Enhanced Base & 61.53 & 88.00 & \textbf{86.17} & 57.70 & 66.73 & \textbf{77.30} & 19.10 & 62.23 & 43.80 & 44.37 & 65.23 & 61.11  \\
        & Enhanced Base + \ours & \textbf{61.90} & \textbf{88.80} & 83.60 & \textbf{59.13} & \textbf{79.17} & 74.03 & \textbf{21.40} & \textbf{62.33} & \textbf{50.20} & \textbf{61.70} & \textbf{64.77} & \textbf{64.28}  \\
        \midrule
        \multirow{4}{*}{2-shot} & Regular Base & 60.33 & 86.29 & \textbf{85.77} & 55.61 & 66.14 & \textbf{77.31} & 17.28 & 58.52 & 42.32 & 37.56 & 61.46 & 58.96   \\
        & Regular Base + \ours & \textbf{62.17} & \textbf{90.13} & 84.43 & \textbf{62.77} & \textbf{85.63} & 75.30 & \textbf{23.07} & \textbf{64.33} & \textbf{54.53} & \textbf{65.77} & \textbf{69.10} & \textbf{67.02}  \\
        \cmidrule(r){2-14}
        & Enhanced Base & 61.87 & 89.37 & \textbf{86.93} & 59.75 & 68.23 & \textbf{77.53} & 19.87 & 63.83 & 46.53 & 49.5 & 67.63 & 62.82  \\
        & Enhanced Base + \ours & \textbf{62.63} & \textbf{90.27} & 84.63 & \textbf{63.70} & \textbf{86.57} & 75.17 & \textbf{24.13} & \textbf{64.97} & \textbf{55.13} & \textbf{65.83} & \textbf{70.00} & \textbf{67.55}  \\
        \midrule
        \multirow{4}{*}{4-shot} & Regular Base & 60.33 & 86.29 & 85.77 & 55.61 & 66.14 & \textbf{77.31} & 17.28 & 58.52 & 42.32 & 37.56 & 61.46 & 58.96   \\
        & Regular Base + \ours & \textbf{62.93} & \textbf{90.63} & \textbf{86.27} & \textbf{66.50} & \textbf{89.50} & 76.23 & \textbf{24.83} & \textbf{66.67} & \textbf{59.50} & \textbf{72.97} & \textbf{69.70} & \textbf{69.61} \\
        \cmidrule(r){2-14}
        & Enhanced Base & 62.43 & 90.33 & \textbf{87.47} & 61.87 & 73.03 & \textbf{77.97} & 20.93 & 65.80 & 49.80 & 49.43 & 69.80 & 64.44  \\
        & Enhanced Base + \ours & \textbf{63.57} & \textbf{90.97} & 86.33 & \textbf{67.43} & \textbf{90.20} & 76.10 & \textbf{25.70} & \textbf{67.27} & \textbf{60.70} & \textbf{73.83} & \textbf{70.93} & \textbf{70.28}  \\
        \midrule
        \multirow{4}{*}{8-shot} & Regular Base & 60.33 & 86.29 & 85.77 & 55.61 & 66.14 & \textbf{77.31} & 17.28 & 58.52 & 42.32 & 37.56 & 61.46 & 58.96  \\
        & Regular Base + \ours & \textbf{64.03} & \textbf{92.23} & \textbf{87.07} & \textbf{70.57} & \textbf{94.30} & 76.90 & \textbf{29.50} & \textbf{68.70} & \textbf{64.23} & \textbf{78.07} & \textbf{74.77} & \textbf{72.76} \\
        \cmidrule(r){2-14}
        & Enhanced Base & 63.33 & 91.60 & \textbf{88.07} & 66.73 & 87.67 & \textbf{78.23} & 23.67 & 68.07 & 59.73 & 67.63 & 74.27 & 69.91 \\
        & Enhanced Base + \ours & \textbf{64.67} & \textbf{92.40} & 87.17 & \textbf{71.83} & \textbf{94.73} & 76.40 & \textbf{31.50} & \textbf{68.73} & \textbf{64.77} & \textbf{79.33} & \textbf{75.33} & \textbf{73.35}  \\
        \midrule
        \multirow{4}{*}{16-shot} & Regular Base & 60.33 & 86.29 & 85.77 & 55.61 & 66.14 & 77.31 & 17.28 & 58.52 & 42.32 & 37.56 & 61.46 & 58.96  \\
        & Regular Base + \ours & \textbf{64.75} & \textbf{92.90} & \textbf{88.10} & \textbf{74.93} & \textbf{96.10} & \textbf{78.23} & \textbf{33.73} & \textbf{70.30} & \textbf{67.57} & \textbf{82.57} & \textbf{76.87} & \textbf{75.10} \\
        \cmidrule(r){2-14}
        & Enhanced Base & 64.13 & 92.57 & \textbf{89.07} & 71.67 & 92.00 & \textbf{78.70} & 27.20 & 70.27 & 64.13 & 76.83 & 77.37 & 73.09 \\
        & Enhanced Base + \ours & \textbf{65.73} & \textbf{93.43} & 87.83 & \textbf{76.83} & \textbf{96.03} & 77.60 & \textbf{36.30} & \textbf{70.67} & \textbf{67.13} & \textbf{84.03} & \textbf{77.97} & \textbf{75.78}  \\
        \bottomrule
    \end{tabular}
    \end{adjustbox}
    \caption{Full numerical results of ablation study of our \ours~effectiveness.
    }
    \label{tab:taskres_effect}
\end{table*}

\paragraph{Ablation study of scaling factor.}
We show the full comparison results across 11 datasets in Table \ref{tab:numerical_alpha}. Generally, our method is not very sensitive to scaling factor $\alpha$ when $\alpha \in [0.3, 1]$, and our \ours~with even $\alpha=0.1$ can also be a strong performance booster (2.90\% accuracy gain). On average, setting $\alpha$ to 0.5 achieves good performance. 
However, the best scaling factor $\alpha$ for various datasets can be different. For instance, a larger $\alpha$ performs better on Flower102 and EuroSAT, while a smaller one is better for OxfordPets and Food101.
\tcb{We then use a learnable parameter (incorporating a $\tanh$ activation) to adaptively determine the value of $\alpha$. On average, the learned $\alpha$ attains the most favorable result.}

\begin{table*}[h]
    \centering
    \begin{adjustbox}{max width = 1.0\textwidth}
    \begin{tabular}{lcccccccccccc}
        \toprule
        $\alpha$ & ImageNet & Caltech101 & OxfordPets & StanfordCars & Flowers102 & Food101 & FGVCAircraft & SUN397 & DTD & EuroSAT & UCF101 & Average \\
        \midrule
        0  & 60.33 & 86.29 & \textbf{85.77} & 55.61 & 66.14 & \textbf{77.31} & 17.28 & 58.52 & 42.32 & 37.56 & 61.46 & 58.96  \\
        0.1 &  60.77 & 87.43 & 84.93 & 59.40 & 70.27 & 75.60 & 19.20 & 59.63 & 46.53 & 54.47 & 62.27 & 61.86 \\
        0.3 &  61.37 & 88.63 & 84.27 & \textbf{59.83} & 75.47 & 74.73 & 20.80 & 61.83 & 49.83 & 60.07 & 64.53 & 63.76 \\
        0.5 &  \textbf{61.43}  & \textbf{88.80}  & 83.50  & 58.77  & 78.00  & 74.03  & 21.20  & \textbf{61.93}  & \textbf{50.17}  & 61.27  & \textbf{64.57}  & \textbf{63.97} \\
        0.7 &  \textbf{61.43} & 88.70 & 82.80 & 57.80 & \textbf{78.90} & 73.17 & \textbf{21.23} & 61.37 & 49.57 & 61.47 & 63.93 & 63.67 \\
        1 &  61.23 & 88.53 & 81.60 & 56.23 & 78.77 & 71.67 & 20.83 & 60.50 & 49.03 & \textbf{61.77} & 63.07 & 63.02 \\
        \midrule
        Learned &  61.33 & 88.73 & 84.00 & 59.47 & 77.40 & 74.40 & 20.63 & \textbf{61.93} & 49.90 & 60.43 & \textbf{65.93} & \textbf{64.01} \\
        \bottomrule
    \end{tabular}
    \end{adjustbox}
    \caption{Full numerical results of ablation study of scaling factor $\alpha$ \tcb{on 1-shot ImageNet}.
    }
    \label{tab:numerical_alpha}
\end{table*}

\begin{table}[h]
\centering
\begin{adjustbox}{max width=0.45\textwidth}
    \begin{tabular}{lccccc}
        \toprule
             Setting  & 1-shot & 2-shot & 4-shot & 8-shot & 16-shot  \\
             \midrule
             Mean  & 0.0124 & 0.0130 & 0.0118 & 0.0200 & 0.0232\\
             Median & 0.0474 & 0.0493 & 0.0519 & 0.0672 & 0.0638\\
        \bottomrule
    \end{tabular}
\end{adjustbox}
    \caption{Mean and Median of learned task residual magnitudes across 11 datasets.}
    \label{tab:magnitude}
\end{table}

\begin{figure*}[t!]
\centering
\begin{adjustbox}{minipage=\textwidth,scale=0.95}
\begin{subfigure}{0.47\textwidth}
    \includegraphics[width=\textwidth]{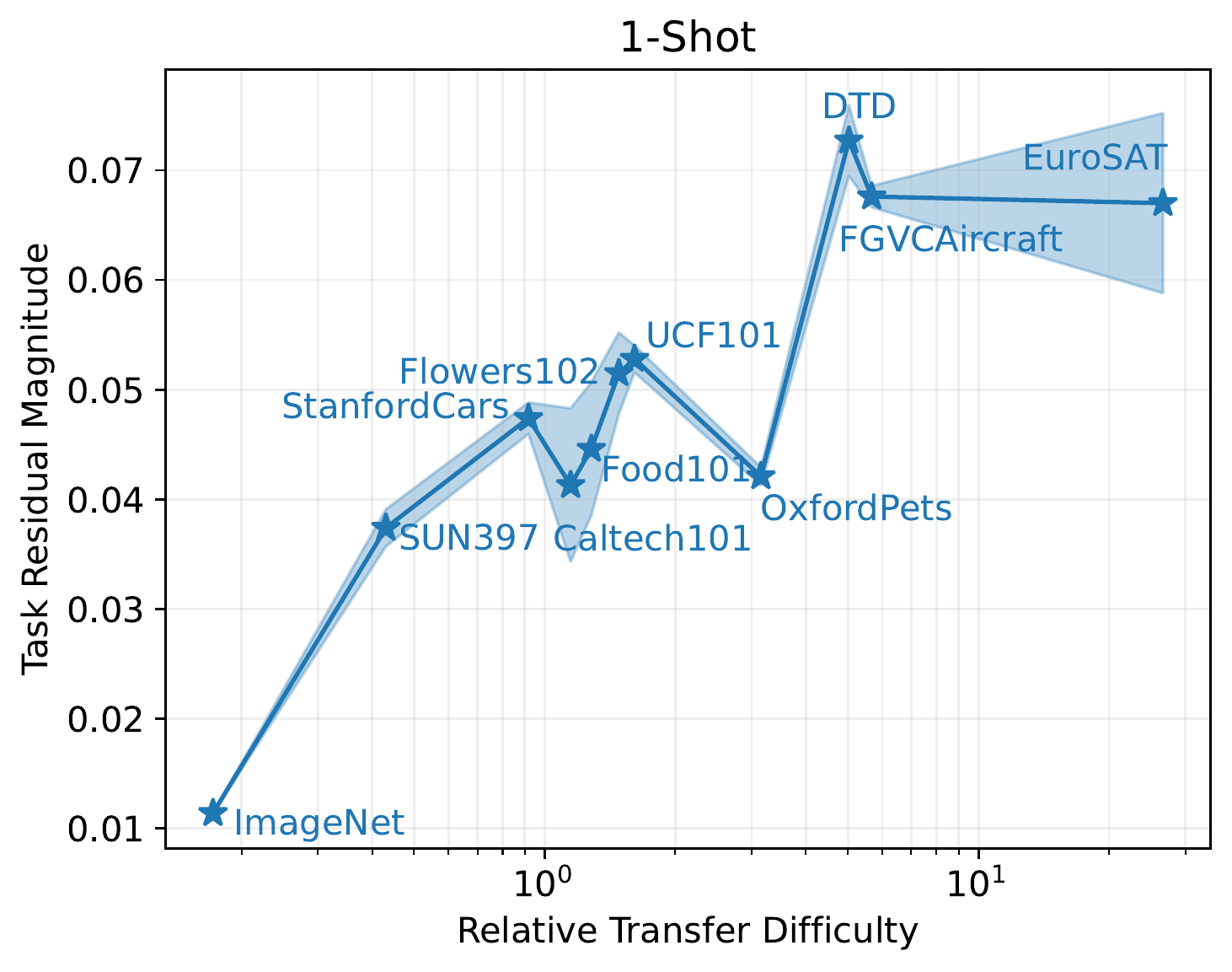}
    \vspace{-10pt}
\end{subfigure}
\hfill
\begin{subfigure}{0.47\textwidth}
    \includegraphics[width=\textwidth]{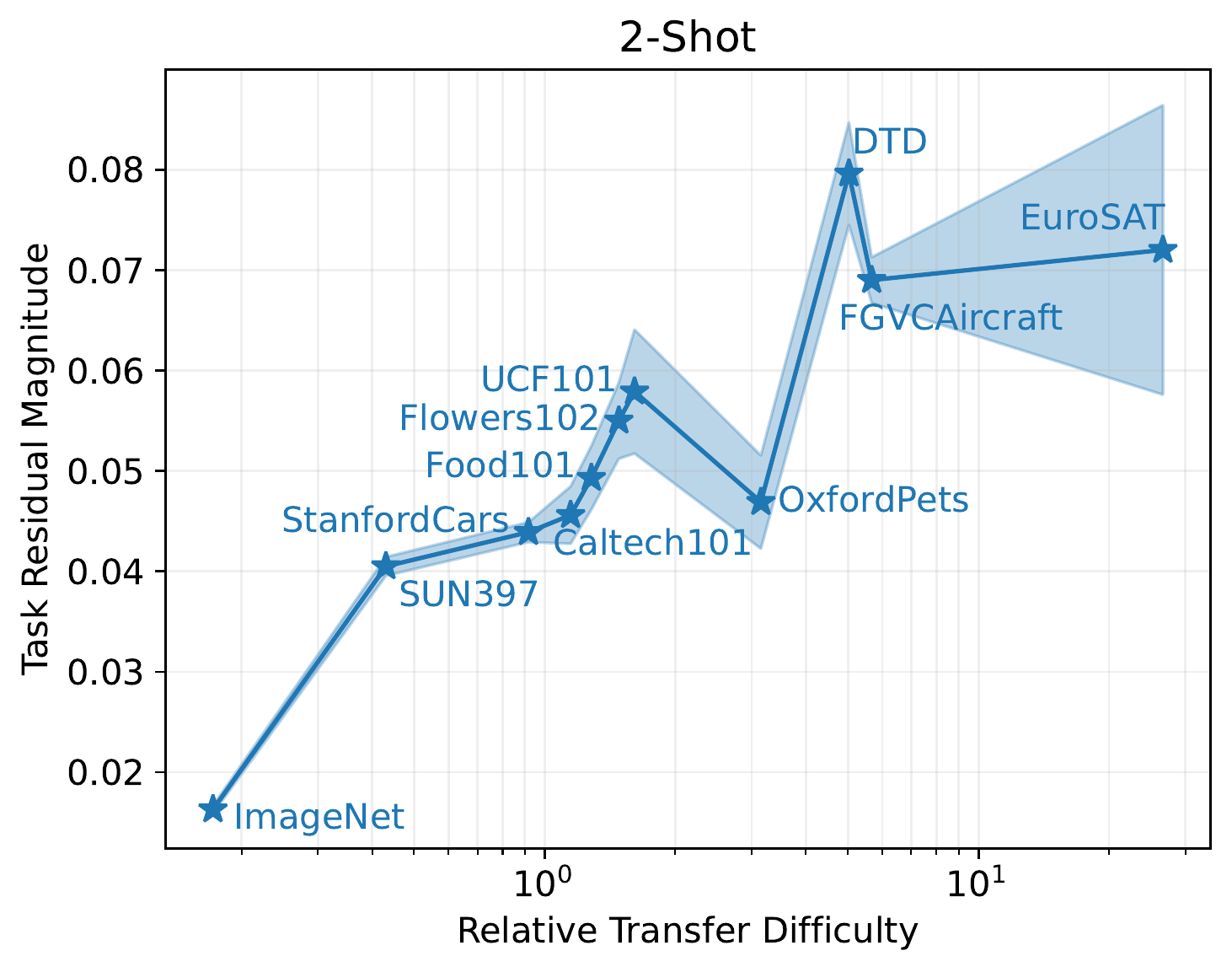}
    \vspace{-10pt}
\end{subfigure}
\hfill
\begin{subfigure}{0.47\textwidth}
    \includegraphics[width=\textwidth]{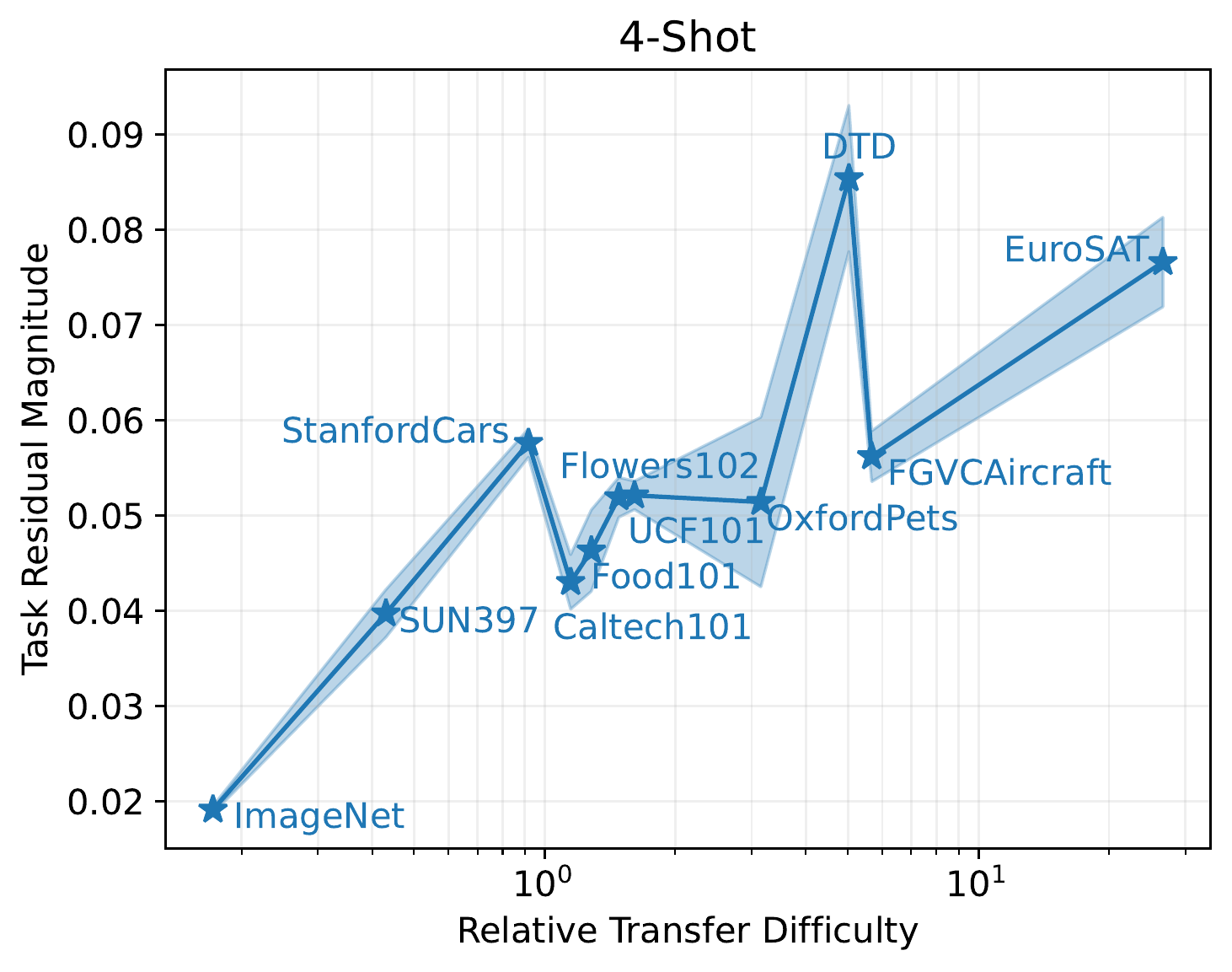}
    \vspace{-10pt}
\end{subfigure}
\hfill
\begin{subfigure}{0.47\textwidth}
    \includegraphics[width=\textwidth]{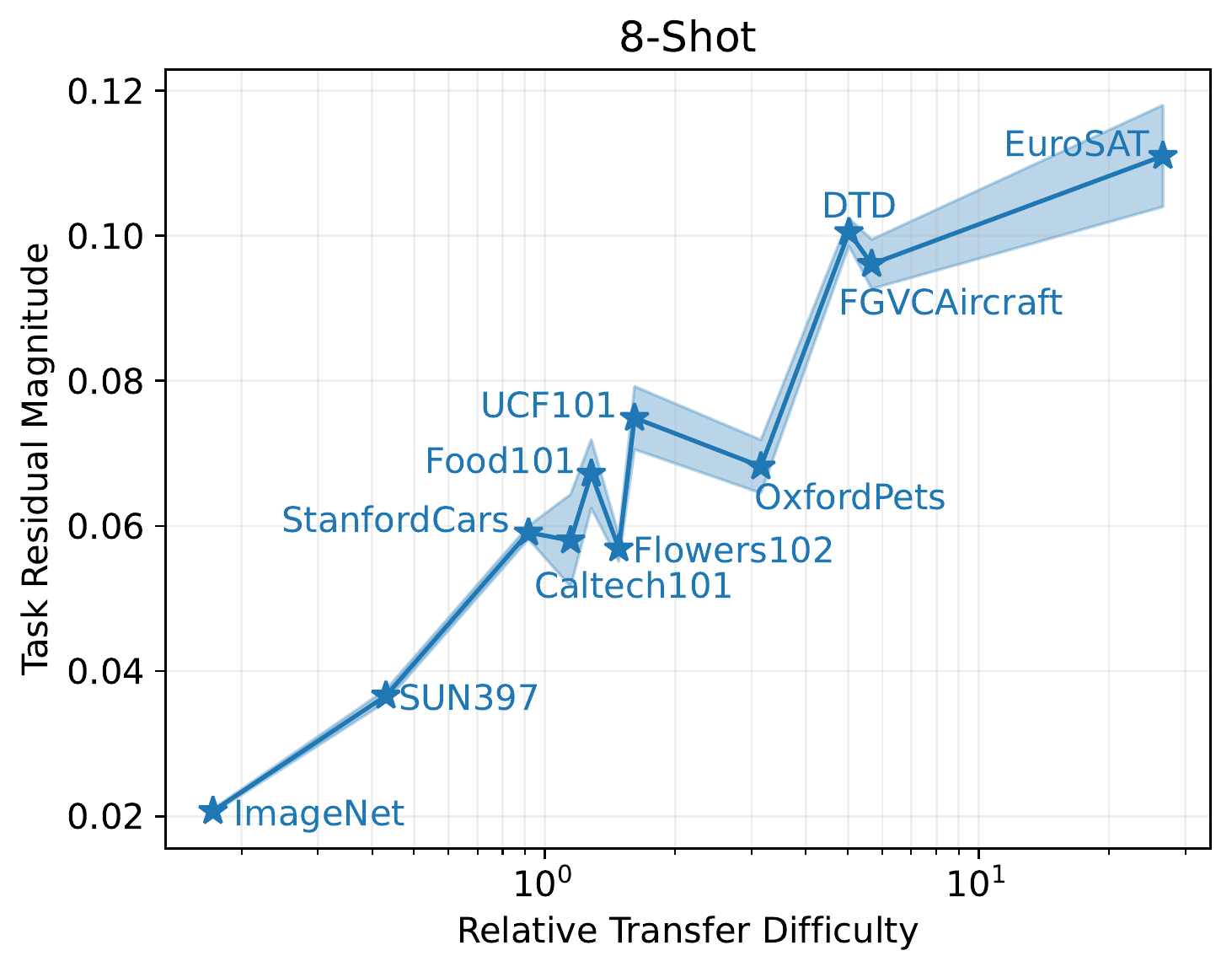}
    \vspace{-10pt}
\end{subfigure}
\end{adjustbox}
\caption{Relation between the magnitude of learned task residuals and the relative transfer difficulty regarding CLIP with 1-/2-/4-/8-shot settings (the 16-shot result is in the main text). The shadow indicates the standard deviation regarding random seeds.}
\label{fig:magnitude}
\end{figure*}

\subsection{Learned Task Residual}\label{apx_vis}
\paragraph{\tcb{More visualization results.}}
We show more results (1-/2-/4-/8-shot settings) of the correlation of learned task residual magnitude and relative transfer difficulty in Figure \ref{fig:magnitude}, and the relation between the learned task residual magnitude and the number of shots in Table \ref{tab:magnitude}. We have the following observations:
\begin{itemize}
    \item The magnitudes of the learned task residuals are positively correlated to the relative transfer difficulty of CLIP for all shot settings, as shown in Figure \ref{fig:magnitude} in this appendix and Figure \ref{fig:taskres_diff} in the main text.
    This shows that the proposed task residual can effectively ``supplement'' the old knowledge according to the task difficulty.
    \item With shot increasing, the mean and median of the learned task residual magnitudes across 11 datasets tend to increase, which indicates that when more downstream task samples are used, more task-specific knowledge can be explored in our method.
    \item With more shots, task-specific knowledge can be captured with less variance as the shadows of the lines are shrinking.
\end{itemize}

\paragraph{\tcb{Does TaskRes effectively preserve the pre-trained boundary?}}
\tcb{To gain deeper insights to the proposed TaskRes, we compare CoOp \cite{zhou2022learning}, CLIP-Adapter \cite{gao2021clip} and TaskRes regarding the number of ``Wrong2Right'' (W2R) images (\ieno, those initially misclassified but later corrected) and ``Right2Wrong'' (R2W) images (\ieno, those initially correctly classified but later misclassified). The models are trained on 4-shot ImageNet and tested on the complete 50k ImageNet test images. 
The W2R/R2W results for the three methods are as follows: (CoOp) 4161/4599, (CLIP-Adapter) 3542/2925, and (TaskRes) 3037/1702. This demonstrates that our TaskRes approach is more effective at preserving the pre-trained decision boundaries compared to other methods. Furthermore, we investigate the commonness of the W2R and R2W images and find that these images tend to occur in the visual concepts sharing the similar high-level semantics, \egno, upright \textit{piano} and grand \textit{piano}.}

\section{Discussion} \label{apx:discussion}
\tcr{\paragraph{Relationship between CLIP-Adapter and our TaskRes.} 
To make the comparison between CLIP-Adapter \cite{gao2021clip} and TaskRes clearer, we here focus on CLIP-Adapter performing on the text branch of CLIP. Given the pre-trained text embeddings $\mathbf{t}$ (\ieno, the text-based classifier), CLIP-Adapter first uses two linear layers $\mathbf{W}_1$ and $\mathbf{W}_2$ (incorporating a ReLU activation) to transform $\mathbf{t}$, and then adds the transformed features to the original embeddings $\mathbf{t}$ to obtain a new classifier. The transformation process (or adapter) can be written as
\begin{equation}
\phi(\mathbf{t})=\text{ReLU}(\mathbf{t}^{T}\mathbf{W}_{1})\mathbf{W}_{2}.
\end{equation}
We can observe that the transformation in CLIP-Adapter has no additive bias, which makes the task-specific learning completely dependent on the old features. 
In contrast, TaskRes introduces a learnable bias $\mathbf{x}$ (\ieno, task residual) that is not relied on the old features (Eq. \ref{eq:target cls} in the main text). This allows for more flexibility in learning task-specific knowledge, leading to better performance.}

\tcr{To further analyze, we extend CLIP-Adapter to two linear transformation versions: linear adapters with and without learnable bias.
Experimental results on 4-shot ImageNet show that linear adapters, both with and without bias (Acc.: 60.93\% and 60.90\%, respectively), underperform the original nonlinear adapter (Acc.: 61.27\%), while the nonlinear adapter is outperformed by our TaskRes (Acc.: 62.93\%). 
This indicates that the key for success is not the use of linear or nonlinear adapters, but the utilization of the \textit{prior-independent} learnable parameters, \ieno, the learnable parameters decoupled from the pre-trained features.}

\tcr{Lastly, although TaskRes could theoretically be considered as a special case of general adapter-style tuning (with adapter $\phi_{\omega}$ parameterized by $\omega$), we believe that the more simplified design and the much stronger performance exhibited by TaskRes have the potential to inspire the community.}

\vspace{-5pt}
\paragraph{\ours~\textit{versus} Tip-Adapter(-F).}
Tip-Adapter \cite{zhang2022tip}, one of the state-of-the-art methods, has a training-free version (\ieno, Tip-Adapter) and an enhanced version Tip-Adapter-F which requires training. Our \ours~has the following differences from Tip-Adapter(-F). (i) Different perspectives: Tip-Adapter(-F) is designed to adjust the classification results (\ieno, logits) produced by the pre-trained classifier via feature retrieval/matching in the training set, while \ours~performs on the weights of the classifier by tuning a prior-independent parameters (\ieno, task residual) added to the pre-trained classifier. Despite the various perspectives, Tip-Adapter(-F) and our \ours~are theoretically complementary.
(ii) Different scalability: The number of tunable parameters of Tip-Adapter-F linearly increases with shot number while ours does not increase, which makes \ours~more scalable than Tip-Adapter-F. While (training-free) Tip-Adapter does not need tunable parameters, the inference of an image requires all training sample features. Besides, the performance of Tip-Adapter largely underperforms Tip-Adapter-F and \ours.

\tcb{\paragraph{Difference between prompt tuning in GLIP and our TaskRes.}
The prompt tuning in GLIP \cite{li2022grounded} performs on the intermediate features $P^0$, which are the outputs of the text encoder and the inputs for subsequent neural networks (NNs) such as BERT layers \cite{kenton2019bert}. During tuning, GLIP omits the text encoder, removing the need to run it at every training step, which is similar to our TaskRes. However, the $P^0$ is subsequently fed into the following NNs, and updating $P^0$ still requires running the NNs (both forward and backward) each time. As a result, GLIP's prompt tuning tends to follow a prompt tuning style.}

\paragraph{\tcb{For which types of tasks does TaskRes yield greater improvements?}}
Our \ours~achieves more significant improvement on tasks where more specialized/expertise knowledge is needed, \egno, EuroSAT, DTD and Flowers102. With 1-shot data, \ours~improves those tasks by \textbf{7.85\% $\sim$ 23.71\%}. With 16-shot data, the improvements are enlarged to \textbf{25.25\% $\sim$ 45.01\%}. This is because our \ours~can effectively learn task-specific knowledge.

\section{Broader Impact}
\tcb{In this work, we conduct experiments and perform analyses based on CLIP \cite{radford2021learning}. However, our proposed concept of learning addictive residual weights for efficient transfer learning is generic and can be adopted to a wider range of vision-language models, such as ALIGN \cite{jia2021scaling}, Perceiver IO \cite{jaegle2022perceiver}, Flamingo \cite{alayrac2022flamingo}, and others. Furthermore, this concept can potentially be extended to tuning vision \cite{he2016deep,dosovitskiy2021an,liu2021swin} or language \cite{kenton2019bert,liu2019roberta} models.



\end{document}